\documentclass[letterpaper, 10 pt, conference]{IEEEtran}  

\IEEEoverridecommandlockouts                              
\usepackage{times}
\usepackage[usenames, dvipsnames]{xcolor}
\usepackage[numbers]{natbib}
\usepackage{multicol}
\usepackage[linkcolor=blue,citecolor=blue,urlcolor=blue]{hyperref}
\usepackage{amsmath}
\usepackage{mathtools}   
\usepackage{amsfonts}
\usepackage{tabularx}
\usepackage{booktabs}
\usepackage{etoolbox}
\usepackage[font={small}]{caption}
\usepackage{subcaption}
\usepackage{eufrak}
\usepackage{stackengine}
\usepackage{multirow}
\usepackage[colorinlistoftodos]{todonotes}
\usepackage{flushend}  

\usepackage{graphicx, import}

\usepackage{xspace}
\usepackage{algorithmic}
\usepackage{algorithm}

\renewcommand{\baselinestretch}{0.96}

\newcommand{\eg}{{e.g.}}
\newcommand{\ie}{{i.e.}}

\newcommand{\coco}{CoCo}
\newcommand{\mlopt}{MLOPT}
\newcommand{\reals}{{\mathbf{R}}}

\newcommand{\bin}{\delta}
\newcommand\mydots{\makebox[1em][c]{.\hfil.\hfil.}}   

\title{\LARGE \bf
Learning Mixed-Integer Convex Optimization Strategies for\\ Robot Planning and Control
}

\author{
\parbox{\linewidth}{\centering
      Abhishek Cauligi$^*$, Preston Culbertson$^*$, Bartolomeo Stellato,\\
Dimitris Bertsimas, Mac Schwager, and Marco Pavone
    }%
\thanks{A. Cauligi, M. Schwager, and M. Pavone are with the Department of Aeronautics and Astronautics, Stanford University, Stanford, CA 94305. \{\tt acauligi, schwager, pavone\} {\tt@stanford.edu}.}
\thanks{P. Culbertson is with the Department of Mechanical Engineering, Stanford University, Stanford, CA 94305. \tt pculbertson@stanford.edu.}
\thanks{B. Stellato is with the Department of Operations Research and Financial Engineering, Princeton University, Princeton, NJ 08544. \tt bstellato@princeton.edu.}
\thanks{D. Bertsimas is with the Operations Research Center, MIT, Cambridge, MA 02139. \tt dbertsim@mit.edu.}
\thanks{This work was supported in part by NASA under the NASA Space Technology Research Fellowship Grants NNX16AM78H and 80NSSC18K1180.}
\thanks{$^*$These authors contributed equally to this work.}
}

\begin{document}

\maketitle
\thispagestyle{empty}
\pagestyle{empty}

\begin{abstract}
Mixed-integer convex programming (MICP) has seen significant algorithmic and hardware improvements with several orders of magnitude solve time speedups compared to 25 years ago. 
Despite these advances, MICP has been rarely applied to real-world robotic control because the solution times are still too slow for online applications.
In this work, we present the \coco{} (Combinatorial Offline, Convex Online) framework to solve MICPs arising in robotics at very high speed.
\coco{} encodes the combinatorial part of the optimal solution into a {\em strategy}.
Using data collected from offline problem solutions, we train a multiclass classifier to predict the optimal {\em strategy} given problem-specific parameters such as states or obstacles.
Compared to~\cite{BertsimasStellato2019}, we use task-specific strategies and prune redundant ones to significantly reduce the number of classes the predictor has to select from, thereby greatly improving scalability.
Given the predicted strategy, the control task becomes a small convex optimization problem that we can solve in milliseconds.
Numerical experiments on a cart-pole system with walls, a free-flying space robot, and task-oriented grasps show that our method provides not only 1 to 2 orders of magnitude speedups compared to state-of-the-art solvers but also performance close to the globally optimal MICP solution.
\end{abstract}

\IEEEpeerreviewmaketitle


\section{Introduction}

Mixed-integer convex programming (MICP) is a powerful technique to model robotic tasks such as task and motion planning~\cite{LandryDeitsEtAl2016,RichardsSchouwenaarsEtAl2002,CulbertsonBandyopadhyayEtAl2019}, 
planning for systems with contact~\cite{DeitsKoolenEtAl2019,MarcucciDeitsEtAl2017} and dexterous manipulation~\cite{HauserWangEtAl2018,HoganGrauEtAl2018}.
By modeling discrete control decisions with integer variables, we can, theoretically, solve MICPs to global optimality with branch-and-bound or outer approximation algorithms~\cite{LeeLeyffer2012}. 
However, computing the optimal solutions is, in practice, computationally challenging.

Despite recent advances~\cite{Bixby2012}, MICPs have seen limited application in real-world robotic tasks for the following two reasons:
\begin{itemize}
\item {\em Slow computation times:} it is still challenging to solve MICPs in online settings where real-time computations are crucial. For example, many planning and control tasks require solutions in few milliseconds, while typical solution times range from seconds to minutes~\cite{MarcucciDeitsEtAl2017}.
\item {\em Non-embeddable algorithms:} the best off-the-shelf solvers such as Gurobi~\cite{ios_gurobi2016} or Mosek~\cite{MosekAPS2010}
rely on complex multithreaded implementations that are not suitable for embedded robot platforms.
\end{itemize}
This is why there is still a large gap between modeling and real-world implementation of MICP-based controllers in robotics systems.
To fill this gap, in this paper we exploit the idea that by solving a large number of MICP instances one can generate a large amount of offline data that can be purposefully used to significantly accelerate online solutions.

\begin{figure}[t!]
\centering
\def\svgwidth{\columnwidth}
\includegraphics[width=0.98\columnwidth]{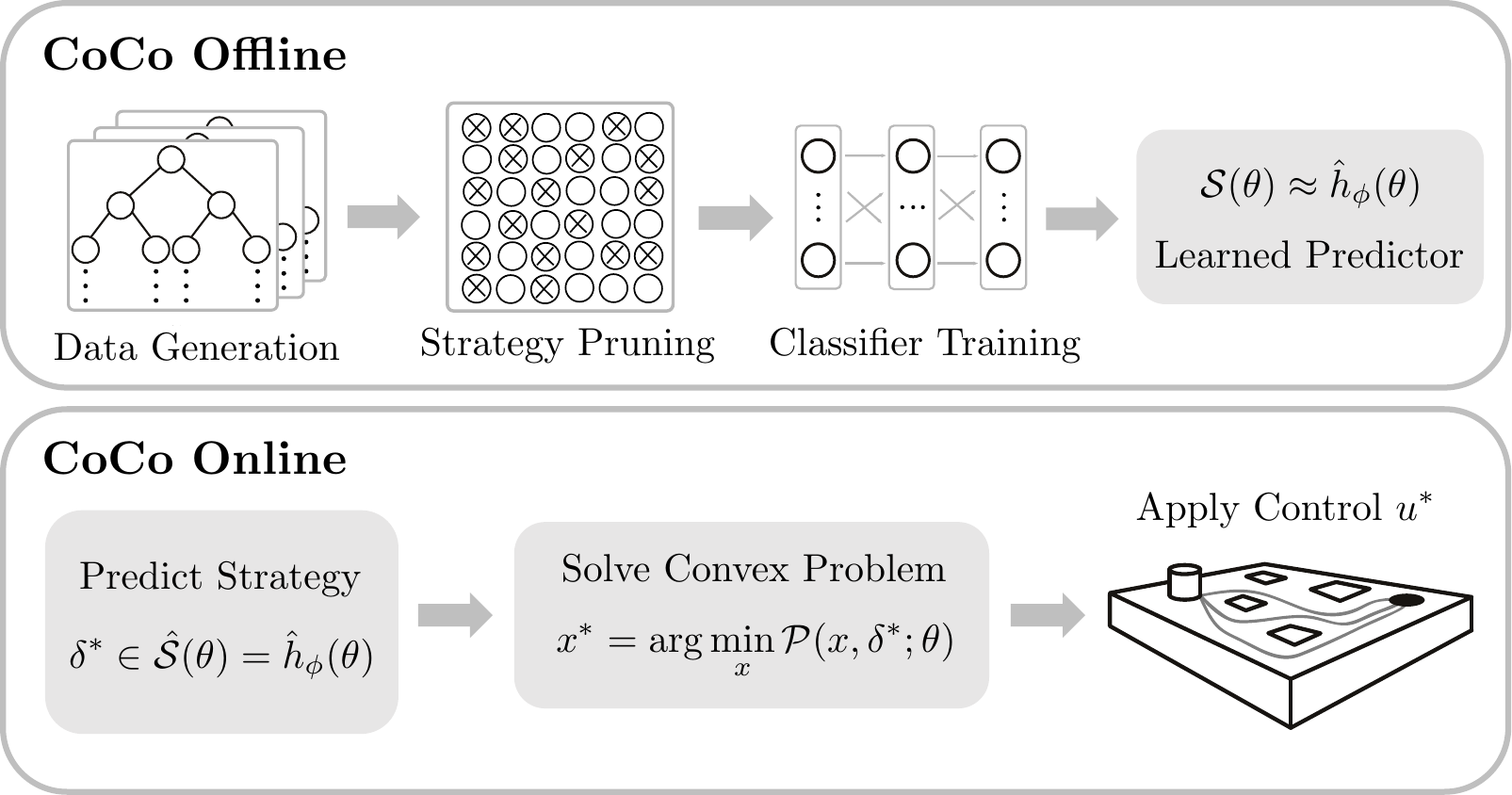}
\caption{\coco{} framework for a motion planning task with free-flyin   g space robots. A strategy classifier is trained offline on a dataset of problems. We added a block to simplify the number of strategies to select. Online, we use the strategy classifier to predict optimal trajectories given a new task specification for the MICP.}
\label{fig:free-flyer_table}
\end{figure}

\subsection{Statement of Contributions}

To leverage offline data to accelerate online solutions, in this paper we introduce Combinatorial Offline, Convex Online (\coco{}), an approach that builds upon the machine learning optimizer (\mlopt{}), recently introduced in~\cite{BertsimasStellato2019,BertsimasStellato2020}. In detail, both \mlopt{} and \coco{} consist of offline and online phases that are notionally represented in Figure~\ref{fig:free-flyer_table}. In the offline phase we collect data by solving a given control problem for several values of the key parameters (\eg{}, initial state, obstacles, object locations), obtaining the optimal solutions.
Given each solution, \mlopt{} saves the optimal integer assignment, which we term an {\em integer} strategy, while \coco{} saves the optimal {\em logical} strategy, both of which we define later.
After the dataset generation we train a neural network which classifies the best strategy to apply given new problem parameters.
Online, the approach consists of a forward neural network strategy prediction from a set of problem parameters, followed by a convex optimization which can be carried out very efficiently~\cite{BoydVandenberghe2004}.
However, \mlopt{} na\"{i}vely applied to robotics problems cannot in general provide reliable time solve times, since the number of unique integer strategies can be too large to obtain high-accuracy multiclass classification.

Here, we demonstrate the efficacy of \coco{}, which leverages problem-specific information and structure to solve MICPs arising in robotics with improved performance and reliability compared to \mlopt{}.
Our detailed contributions are as follows.
First, we propose a methodology to identify unique logical strategies which encompass a set of redundant integer solutions corresponding to the same globally optimal continuous solution.
This happens often in robotic settings, for example, when we obtain multiple integer assignments for the same obstacle avoidance path or multi-contact trajectory.
Second, we introduce the notion of {\em task-specific} strategies that allows us to exploit the separable structure of robotics problems, thereby greatly improving scalability. Examples of structured problems include path planning where we can decouple combinatorial decisions obstacle-wise. For these settings, we learn a predictor for a single combinatorial decision, \eg, on which side of the obstacle the robot should pass. Online, we can apply the same predictor for every combinatorial decision we have to make. This idea greatly simplifies the number of unique strategies encountered by focusing on every independent decision.
Finally, numerical experiments on a cart-pole system with walls, a free-flying space robot, and task-oriented grasps show that our method achieves strategy prediction accuracy which is at least $90\%$ feasible with more than $90\%$ of solutions being globally optimal. In addition, we obtain computational speedups from 1 to 2 orders of magnitude compared to MICP solvers Gurobi~\cite{ios_gurobi2016} and Mosek~\cite{MosekAPS2010}.
Therefore, the proposed algorithm is suitable to compute MICP solutions in real-time reliably and at very high-speed.

\subsection{Related Work}
\label{sec:related_work}
Recently, there has been a surge of interest in applying data-driven methods in accelerating solution times for optimization-based controllers. In~\cite{ChenWangEtAl2019,ZhangBujarbaruahEtAl2019}, neural networks are used to warm start a solver for a QP-based controller used in a receding horizon fashion. Tang, et al. also consider non-convex optimization-based controllers by learning warm starts for an SQP-based trajectory optimization library~\cite{TangSunEtAl2018} and for indirect optimal control methods \cite{TangHauser2017}. Using tools from differentiable convex optimization, a learning-based approach to tune optimization-based controllers is proposed in~\cite{AgrawalBarrattEtAl2019b}. While these methods have shown considerable promise, they only consider the setting of continuous optimization.

Relatively less attention has been paid to using data-driven techniques for accelerating integer program solutions in control~\cite{BengioLodiEtAl2018,LodiZarpellon2017}. A popular class of methods has been to pose branch-and-bound as a sequential decision making process and apply imitation learning to learn effective variable and node selection strategies~\cite{DaiKhalilEtAl2017, HeDaumeIIIEtAl2014}. However, these approaches can still be too computationally expensive for applications in control, as they require solving branch-and-bound and computing multiple forward passes of a neural network online. 

Masti and Bemporad~\cite{MastiBemporad2019} use a regression based approach to train a neural network to learn binary variable assignments. However, the authors only demonstrate this approach on an MIQP with 14 binary variables. An approximate dynamic programming approach is presented in~\cite{DeitsKoolenEtAl2019}, where the cost-to-go function for an MIQP controller is learned. However, the approach still requires solving an expensive MIQP using branch-and-bound online. Hogan, et al.~\cite{HoganGrauEtAl2018} propose an approach similar to ours in that they train a classifier to predict mode sequences for a manipulation task, but our formulation is more general in being able to handle a larger class of mixed logical dynamical systems.

Our approach draws inspiration from the field of explicit MPC \cite{DomahidiZeilingerEtAl2011}. However, instead of learning feedback control laws for polyhedral regions of the state space, we learn strategies for regions of the parameter space of the problem. Our work is also inspired by recent methods from multi-task and meta learning \cite{AchilleLamEtAl2019, RakellyZhouEtAl2019} which use \textit{task embeddings} to identify and solve various tasks. From this view, the optimal control problem parameters correspond to a pre-defined task embedding, and our machine learning problem is to map this embedding to a strategy which yields a convex optimization solution with low cost.

\section{Technical Background}
\subsection{Mixed-Integer Convex Programs}
\label{subsec:bips}
In this work, we focus our attention on a specific class of discrete optimization problems known as parametrized mixed-integer convex programs. The general form of this problem is:
\begin{equation} \label{eq:bcp}
\begin{array}{ll}
\text{minimize}\!\!\!&f_0(x,\bin;\theta) \\
\text{subject to}\!\!\!& f_i(x,\bin;\theta) \leq 0, \quad i = 1,\dots,m_f \\
&\bin\in \{0,1\}^{n_\bin},\\
\end{array}
\end{equation}  
where $x\in\reals^{n_x}$ are continuous decision variables, $\bin \in \{0,1\}^{n_\bin}$ are binary decision variables, and $\theta \in \reals^{n_p}$ are the problem parameters. The objective function $f_0$ and inequality constraints $f_i$ are convex. While problem \eqref{eq:bcp} is $\mathcal{NP}$-hard due to the inclusion of discrete decision variables $\bin$~\cite{Karp1975}, it can be solved to global optimality using algorithms such as branch-and-bound, which operates by sequentially building an enumeration tree of relaxed problems to the original problem.

\subsection{Strategies for MICPs}
Here, we briefly review the notion of an integer strategy as defined in~\cite{BertsimasStellato2019,BertsimasStellato2020}. For an MICP defined by parameters $\theta \in \reals^{n_p}$, an {\em integer} strategy $\mathcal{I} (\theta)$ consists of a tuple $\big(\bin^*,\mathcal{T}(\theta)\big)$ where ($x^*$, $\bin^*$) is an optimizer of the problem (\ref{eq:bcp}), and $\mathcal{T}(\theta) = \{ i \in \left\{1, \ldots, m_f\right\} \mid f_i(x^*, \delta^*;\theta) = 0 \}$ is the corresponding set of active inequality constraints.

Given an optimal integer strategy $\mathcal{I} (\theta)$, we can obtain an optimal solution for \eqref{eq:bcp} by solving a convex optimization problem,
\begin{equation*}
\begin{array}{ll}
\text{minimize} & f_0(x,\bin^*;\theta) \\
\text{subject to} & f_i(x,\bin^*;\theta) \leq 0, \:\: i \in \mathcal{T}(\theta),\\
\end{array}
\end{equation*}
which is much easier than the original MICP (\ref{eq:bcp}). In fact, if the original problem (\ref{eq:bcp}) is a mixed-integer quadratic program (MIQP) or mixed-integer linear program (MILP), solving the reduced problem corresponds to simply solving a set of linear equations defined by the KKT conditions.

This insight motivates the supervised learning problem considered in~\cite{BertsimasStellato2020}, wherein the authors aim to learn an approximate mapping $h_\phi$ from problem parameters $\theta$ to a corresponding integer strategy $\mathcal{I} (\theta).$ The authors pose this as a multiclass classification problem over a dataset $\mathcal{D} = \left\{(\theta_i,\mathcal{I}_i)\right\}_{i=1}^T$ of problem parameters $\theta_i$ and their corresponding strategies $\mathcal{I}_i$. Here, the $\theta_i$ are sampled from $p(\Theta)$, which is a (known) parameter distribution representative of the problems encountered in practice.

\subsection{Big-M Formulations of Mixed Logical Dynamical Systems}
A common modeling choice in MICPs is to use the binary variables $\bin$ to capture high-level discrete or logical behavior of the system (\eg, contact, task assignment, hybrid control logic), and to enforce the resulting high-level behavior on the continuous variables $x$ using what is known as big-M formulation~\cite{BemporadMorari1999}.

For example, suppose we have a logical variable $\bin_i$, which determines if the constraint $g_i(x;\theta) \leq 0$ on the continuous variables $x$ is active, \ie, we seek to impose the relation
\begin{equation}
    \left[\bin_i = 1 \right] \implies \left[g_i(x;\theta) \leq 0\right].
    \label{eq:impl-constraint}
\end{equation}

To this end, let us define
\begin{equation*}
    M_i(\theta) = \sup_{x} g_i(x; \theta),
\end{equation*}
which is simply an upper bound on the constraint function, and may be precomputed. Then, \eqref{eq:impl-constraint} can be enforced by imposing the constraint
\begin{equation}
    g_i(x;\theta) \leq M_i(\theta)(1-\bin_i),
    \label{eq:eg_big_m}
\end{equation}
which we note is linear in $\bin_i$. By inspection, when $\bin_i = 1$, we must have $g_i(x;\theta) \leq 0$, and when $\bin_i = 0$, we have the trivial constraint $g_i(x;\theta) \leq M_i(\theta)$. More complex logical behavior (such as disjunctive constraints) can be achieved using additional logical variables and constraints; for a more thorough treatment, we refer the reader to~\cite{BemporadMorari1999}.

More generally, let us denote a ``big-M'' constraint as
\begin{equation*}
    g_i(x; \theta) \leq a_i(\theta) \bin_i, 
\end{equation*}
where $a_i(\theta)$ are constants which, in general, bounds on the constraint functions to impose the desired logical behavior. 

\subsection{Uniqueness of Global Optima}
We further note that since MICPs are non-convex, they may admit multiple global optima. Yet, for many physical systems, while the program itself is non-convex, the continuous optimizers $x^*$, (\eg, the shortest path through an obstacle field) are generally unique. Thus, we say a program is \textit{well-posed} if it admits a unique continuous minimizer $x^*$, and we say a program is \textit{completely well-posed} if it admits a unique global minimizer $(x^*,\bin^*).$ In this work, we assume the problems considered are well-posed, although they may admit multiple discrete optimizers $\bin^*$.

To this end, for a particular solution $(x^*, \bin^*)$, we consider a big-M constraint \textit{tight} if
\begin{equation*}
    g_i(x; \theta) \leq a_i(\theta) \bin_i  \iff \bin_i = \bin_i^* ,
\end{equation*}
meaning that for the particular continuous solution $x^*$, this constraint may only be satisfied by the values from solution $\bin^*$. For example, considering the implication constraint \eqref{eq:eg_big_m}, suppose for some particular solution $x^*$, $g_i(x^*; \theta) \leq 0$. Then, since the inequality may be satisfied by either $\bin_i = 0$ or $\bin_i = 1$, $\bin_i^*$ is not unique; thus, this big-M constraint is not tight.

In this work, we use big-M constraints exclusively to relate the continuous variables $x$ and logical variables $\bin$. If $m_M$ is the number of big-M constraints used, we can write more specifically the class of MICPs studied:
\begin{equation} \label{eq:bcp_logic}
\!\begin{array}{ll}
\text{minimize} & f_0(x,\bin;\theta) \\
\text{subject to} & f_i(x;\theta) \leq 0, \quad i = 1,\dots,m_f \\
& g_i(x; \theta) \leq a_i(\theta) \bin_i, \quad i = 1,\dots,m_M\\
&\bin\in \{0,1\}^{n_\bin}\\
\end{array}
\tag{$\mathcal{P}(\theta)$}
\end{equation}

\section{Supervised Learning Strategies for MICPs}
\label{sec:supervised_learning_micps}
In this section we show how to apply \coco{} to solve online MICP control tasks through the use of logical strategies and by designing task-specific strategies exploiting the structure of the problem. We finally introduce the complete algorithm and discuss online and offline details.

\subsection{Pruning Redundant Strategies}
\label{subsec:pruning_strategies}
Although the strategy classification problem presented in~\cite{BertsimasStellato2020} provided promising results, it is unable to handle the general class of MICP problems commonly appearing in robotics, and specifically those using integer variables to model mixed logical dynamical systems~\cite{BemporadMorari1999}. A specific pitfall with na{\"i}vely using the integer strategies $\mathcal{I} (\theta)$ from~\cite{BertsimasStellato2020} occurs when considering a problem which is well-posed (\ie, it has a unique optimal value and continuous optimizer $x^*$), but admits multiple discrete optimizers $\{\bin^*\}$. In this case, there exist multiple optimal strategies $\{(\bin^*, \mathcal{T}(\theta))\}$, and thus multiple correct labels for the same problem data $\theta$, which makes the supervised learning problem ill-posed. This can occur, for example, when implication constraints \eqref{eq:impl-constraint} are included in the model, or when using logical ``OR'' constraints between variables.

To ameliorate this issue, we leverage the  insight that for our problem~\ref{eq:bcp_logic}, a strategy $\mathcal{S}(\theta)$ can be equivalently, and uniquely, defined by considering only the set of big-M constraints which are tight.

Thus, let us instead define a logical strategy $\mathcal{S}(\theta)$ as a tuple $(\bin^*(\theta), \mathcal{T}_M(\theta))$, where $\bin^*(\theta)$ is a particular integer solution, and $\mathcal{T}_M(\theta)$ is the set of tight big-M constraints,
\begin{align}
\mathcal{T}_M(\theta) = \{ i \mid g_i(x^*;\theta) \leq a_i(\theta)\bin_i \iff \bin_i = \bin_i^* \}. \label{eq:strategy_defn}
\end{align}
Since each $\bin_i \in \left\{0,1\right\}$, this set may be easily computed. Thus, by definition, our classification problem is once again well-posed.

While we could additionally label which continuous constraints are active, as in~\cite{BertsimasStellato2019}, we note that many common problems in robotics are more general than MIQPs or MILPs, and thus cannot be solved via the KKT conditions. Instead, at runtime, we choose to solve the corresponding convex optimization which results from simply setting $y = \bin^*$.

\begin{figure}[h]
\centering
\def\svgwidth{0.7\columnwidth}
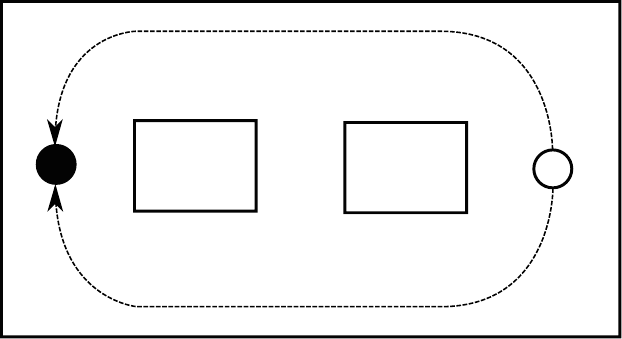
\caption{A motion planning problem where two continuous solutions traversing either side of an obstacle have the same optimal objective.}
\label{fig:measure-zero_strategies}
\end{figure}

Although this revised notion of strategies for MICPs alleviates the concerns of multiple integer assignments $\bin^*$, in the case of an ill-posed problem, there may also exist multiple optimal continuous solutions $x^*$ which achieve the globally optimal objective. For example, Figure \ref{fig:measure-zero_strategies} illustrates a case in which multiple $x^*$ correspond to two different homotopy classes for a motion planning problem around a symmetric set of obstacles. In this work, we neglect such cases, since they are somewhat rare for physical problems, and usually correspond to such pathological cases.

\subsection{Task-Specific Strategies}
\label{subsec:task-specific_strategies}
Extending the notion of the strategy defined in \eqref{eq:strategy_defn}, we can refine the definition of a strategy to exploit problem structure commonly found in many robotics problems --- separability. In physical planning problems, it frequently occurs that sets of logical variables are used to model repeated, but spatially or temporally distinct, phenomena. For example, in the obstacle avoidance problem outlined in~\cite{RichardsSchouwenaarsEtAl2002}, a set of $4$ logical variables is used to encode each obstacle; except through the optimal continuous variables, these logical variables are completely decoupled and do not appear together in common constraints. Similar structure can be found in multi-surface contact planning (variable sets per surface), piecewise-affine dynamics (variable sets per dynamics region), and so on. 

We formalize this notion by considering problems in which the underlying mixed logical constraints can be represented as a conjunction of Boolean formulas,
\begin{equation*}F_1 \land F_2 \land \mydots \land F_\ell,
\end{equation*}
where $F_i$ is a distinct sub-formula of literals involving continuous and logical variables, and each integer variable $\bin_j$ is associated with only one, sub-formula $F_i$. Again, for illustration, the constraints in \cite{RichardsSchouwenaarsEtAl2002} can be represented in this form, where each sub-formula $F_i$ encodes that the robot position must lie outside of a single obstacle at a single timestep.

Suppose the mixed logical constraints of MICP~\ref{eq:bcp_logic} may be expressed in this manner. Then, the strategy $\mathcal{S}(\theta)$ can be decomposed further into sub-formula strategies $\mathcal{S}_1(\theta), \ldots, \mathcal{S}_\ell(\theta)$, without loss of generality, since trivially the value of variable $\delta_j$ can be determined solely from the logical value of its parent sub-formula $F_i$ and the continuous solution~$x^*$.

The practical advantages of this view are twofold: first, since the number of possible strategies can grow exponentially with the number of integer variables $\lvert \bin \rvert,$ decomposing the strategies in this way results in smaller sub-formulas which take on far fewer values individually than the total problem strategy $\mathcal{S}$. Second, many of the individual formulas are themselves repeated (e.g. multiple obstacles, but of various sizes and locations); decomposing the problem strategy into sub-formulas exposes this structure as well. Taken together, these advantages of strategy decomposition greatly help the strategy classifier proposed below, since now the number of possible classes becomes much lower, and a single classifier can be trained for all sub-formulas of identical structure, which in effect augments the training dataset. 

We further note that this approach has diminishing returns; since, at runtime, a set of sub-formula strategies $\mathcal{S}_i$ must be recombined to create the total problem strategy $\mathcal{S}$, any machine learning-based algorithm must now produce $\ell$ correct predictions to reconstruct the globally optimal strategy $\mathcal{S}.$ Thus, when choosing how to decompose the strategy, there is a practical trade-off between the number of possible assignments per sub-formula, and the number of total sub-formula assignments which must be predicted overall.

\subsection{Algorithm Overview}
\begin{algorithm}[tb!]
\caption{\coco{} Offline}
\label{alg:MLOPT_OFFLINE}
\begin{algorithmic}[1]
{\small
    \REQUIRE {Batch of training data} $\{ \theta_i\}_{i=1,\ldots ,T}${, problem}~\ref{eq:bcp_logic}
    \STATE Initialize strategy dictionary $\mathcal{S} \leftarrow \{ \},$ {train set } $\mathcal{D} \leftarrow \{ \}$
    \STATE $k \leftarrow 0$
    \FOR{{each} $\theta_i$}
        \STATE Solve \ref{eq:bcp_logic} 
        \IF{\ref{eq:bcp_logic} is feasible}
            \STATE Construct optimal strategy $\mathcal{S}^*$
            \IF{$\mathcal{S}^* \textbf{ not in } \mathcal{S}$}
                \STATE Add $\mathcal{S}^*$ to $\mathcal{S}$, assign class $k$
                \STATE $k \gets k + 1$
            \ENDIF
            \STATE Assign training label $y_i$ of strategy class $S^*$
            \STATE Add $\left(\theta_i, y_i\right)$ to $\mathcal{D}$
        \ENDIF
    \ENDFOR
    \STATE Choose network weights $\phi$ which minimize cross-entropy loss $\mathcal{L}({h}_\phi(\theta_i),y_i)_{i=1,\ldots,T}$ via stochastic gradient descent
    \RETURN ${h}_\phi$, $\mathcal{S}$
}
\end{algorithmic}
\end{algorithm}

We now detail our proposed procedure for training and deploying a strategy classifier. The offline portion of \coco{} is described in Algorithm \ref{alg:MLOPT_OFFLINE}. In this offline phase, the optimization problem \ref{eq:bcp_logic}  is solved for a batch of sample points $\theta_i$ sampled from a parameter distribution $p(\Theta)$. We maintain a dictionary $\mathcal{S}$ of logical strategies encountered in the dataset, and label each point $\theta_i$ with the index $y_i$ of its strategy class $\mathcal{S}^*$. Finally, we use stochastic gradient descent to choose neural network weights $\phi$ which approximately minimize a cross-entropy loss over the training set.

\begin{algorithm}[t!]
\caption{\coco{} Online}
\label{alg:MLOPT_ONLINE}
\begin{algorithmic}[1]
{\small
    \REQUIRE Problem parameters $\theta$, strategy dictionary $\mathcal{S}$, trained neural network ${h}_\phi$, $n_{\text{evals}}$
    \STATE Compute class scores ${h}_\phi (\theta)$
    \STATE Identify top $n_{\text{evals}}$-scoring strategies in $\mathcal{S}$
    \FOR {$j = 1,\ldots,n_{\text{evals}}$}
        \IF {\ref{eq:bcp_logic} is feasible for strategy $\mathcal{S}^{(j)}$}
            \RETURN {Feasible solution $(x^*,\bin^*)$}
        \ENDIF
    \ENDFOR
    \RETURN \text{failure}
}
\end{algorithmic}
\end{algorithm}
As detailed in Algorithm \ref{alg:MLOPT_ONLINE}, at runtime, we use the provided $\hat{h}_\phi$ and $\mathcal{S}$ online to solve the MICP given new task parameters. Provided a new task specification $\theta$, we then compute a forward pass $\hat{h}_\phi (\theta)$ and identify the strategies with the highest $n_{\text{evals}}$ scores. For each candidate strategy, we fix the integer variaxbles $\bin$ to an optimal assignment $\bin^*$ for the strategy class. We then solve the resulting convex problem to find optimal continuous values $x^*$. If a feasible solution is found, the algorithm terminates.

\subsection{Feasible Solutions}
\label{subsec:feasible_discuss}
We note that \coco{} forgoes the optimality guarantees provided by branch-and-bound for finding the globally optimal solution for an MICP. However, we emphasize that in the context of robotics, controllers are often deployed in a receding horizon fashion wherein the controller is executed multiple times through a task or trajectory. Thus, some suboptimality in relatively rare instances can be tolerated if fast execution enables safe behavior and, in some cases, a well-tuned terminal cost can also account for long horizon behavior. Moreover, as we show next, in practice \coco{} provides better quality solutions within a handful of strategy evaluations than the equivalent of branch-and-bound terminated after a fixed number of iterations.

\section{Numerical Experiments}
\label{sec:numerical_experiments}
In this section, we numerically validate our approach on three benchmark problems in robotics: the control of an underactuated cart-pole with multiple contacts, the robot motion planning problem, and dexterous manipulation for task-specific grasping. These problems demonstrate the rich set of combinatorial constraints that arise in robotics and demonstrate the broad applicability of \coco{} in handling these constraints. Table \ref{table:systems_overview} provides an overview of the three systems.

\begin{table}[h]
\centering
\small
\begin{tabular}{ @{}r l l l l l@{} }
 \toprule
 System &  $n_x$ & $n_\bin$ & $n_p$ \\
 \midrule
 Free-Flyer & 34 & 160 & 44\\
 Cart-Pole & 74 & 40 & 13\\
 Manipulator & 1092 & 30 & 12\\
 \bottomrule
\end{tabular}
\caption{Overview of system dimensions in numerical experiments.}
\label{table:systems_overview}
\end{table}

\subsection{Implementation Details}
For each system, we first generate a dataset by sampling $\theta$ from $p(\Theta),$ until a sufficient number of problems \ref{eq:bcp_logic} are solved. For each system, we separate 90\% of the problems for training and the remaining 10\% for evaluation. For our neural network architecture, we implemented a standard feedforward network with three layers. We use 32 neurons per layer for the cart-pole and dextrous manipulation examples and 128 neurons per layer for the free-flying space robot example. The ReLU activation function was used for each network.

We implemented each example in this section in a library written in Julia 1.2 and the \verb|PyTorch| machine learning library~\cite{PaszkeGrossEtAl2017} to implement our neural network models with the ADAM optimizer for training. The mixed-integer convex problems were written using the JuMP modeling framework~\cite{DunningHuchetteEtAl2017} and solved using Gurobi~\cite{ios_gurobi2016} and Mosek~\cite{MosekAPS2010}. We further benchmark \coco{} against Gurobi, Mosek, and the regression framework from~\cite{MastiBemporad2019}. For all problems except the grasp optimization, we disable presolve and multithreading to better approximate the computational resources of an embedded processor. The network architecture chosen for the regressor was identical to the strategy classifier, updated with the appropriate number of integer outputs. The code for our algorithm is available at~\url{https://github.com/StanfordASL/CoCo}.




\subsection{Cart-Pole with Soft Walls}
\begin{figure*}[t!]
\centering
    \begin{subfigure}[t]{0.2\textwidth}
  \centering
  \captionsetup{justification=centering}
    \includegraphics[width=1.75in,]{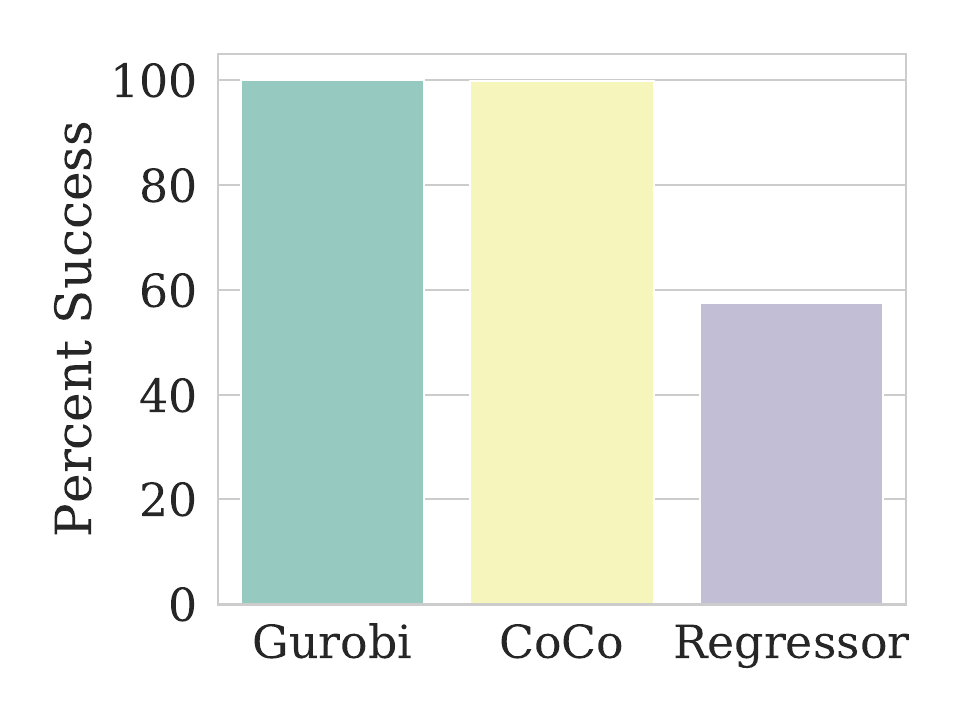}
    \caption{Success percentage}
    \label{fig:cart-pole_percent_succ}
  \end{subfigure}
  \qquad
  \begin{subfigure}[t]{0.2\textwidth}
  \centering
  \captionsetup{justification=centering}
    \includegraphics[width=1.75in]{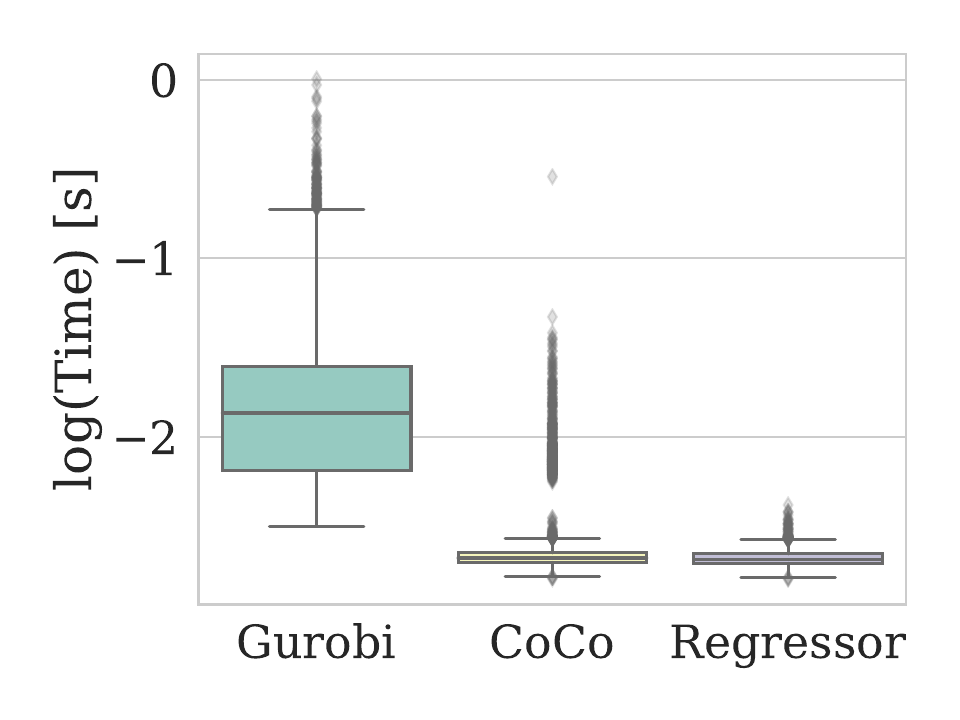}
    \caption{Solution times}
    \label{fig:cart-pole_time}
  \end{subfigure}
  \qquad
  \begin{subfigure}[t]{0.2\textwidth}
    \centering
    \captionsetup{justification=centering}
    \includegraphics[width=1.75in]{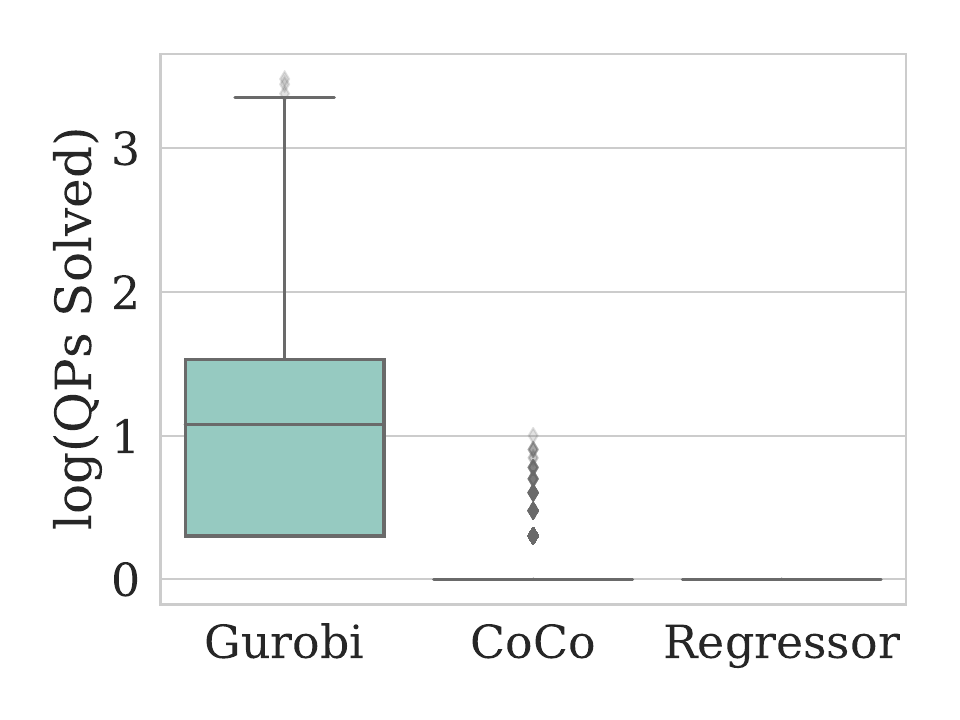}
    \caption{Num. QPs solved}
    \label{fig:cart-pole_solved}
  \end{subfigure}
  \qquad
    \begin{subfigure}[t]{0.2\textwidth}
    \centering
    \captionsetup{justification=centering}
    \includegraphics[width=1.75in]{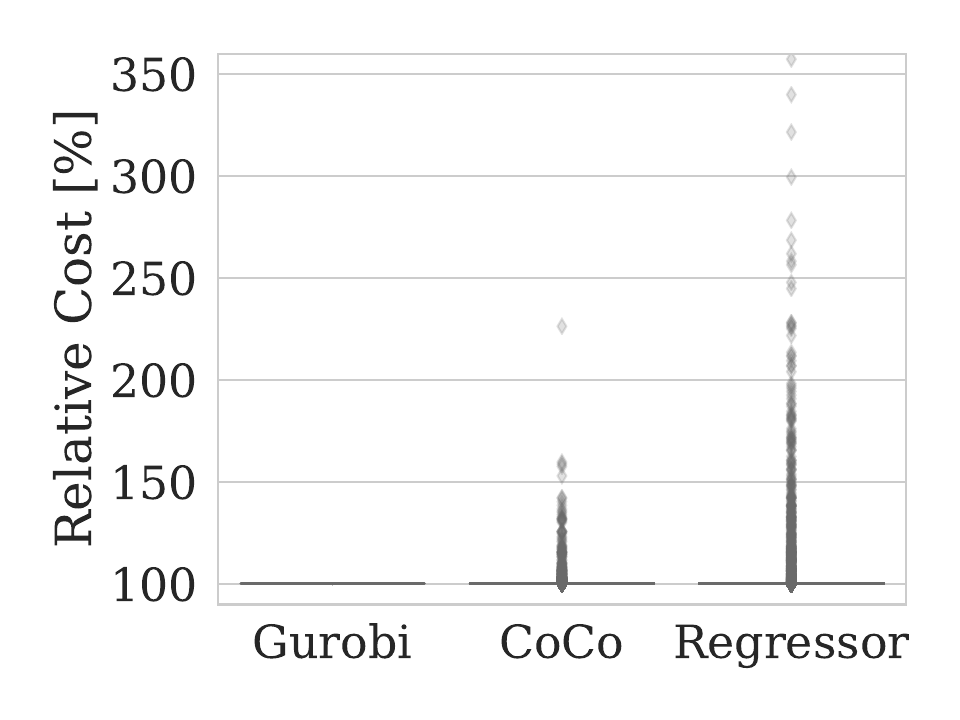}
    \caption{Normalized cost [\%]}
    \label{fig:cart-pole_cost}
  \end{subfigure}
  \caption{Simulation results for the cart-pole system demonstrate the near-total feasibility for solutions of \coco{} without sacrificing optimality.}
\label{fig:cart-pole_results}
\end{figure*}

\begin{figure}[ht]
\centering
\def\svgwidth{0.6\columnwidth}
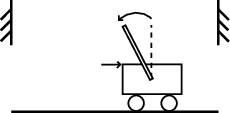
\caption{4D cart-pole with wall system.}
\label{fig:cart-pole_soft_walls}
\end{figure}

For our first system, we consider problems in robotics that deal with planning multi-contact behaviors. While a myriad of approaches have been proposed for planning open-loop plans for systems with contact or discontinuities, developing controllers that can operate online to quickly react to disturbances has proven challenging. Here, we consider the cart-pole with contact system, a well-known underactuated, multi-contact problem in robot control~\cite{DeitsKoolenEtAl2019,MarcucciTedrake2020}. As depicted in Figure \ref{fig:cart-pole_soft_walls}, the system consists of a cart and pole and the optimal control problem entails regulating the system to a goal $x_g$:
\begin{equation} \label{eq:cart-pole_ocp}
\begin{array}{ll}
\text{minimize} & \displaystyle \|x_N-x_g\|_{2} + \sum \limits_{\tau=0}^{N-1} \|x_\tau-x_g\|_{2} + \|u_\tau\|_{2} \\
\text{subject to} & x_{t+1} = Ax_t + Bu_t + G s_t,\quad t = 0,\mydots,N-1 \\
& u_{\textrm{min}} \leq u_t \leq u_{\textrm{max}}, \quad t = 0,\mydots,N-1 \\
& s_t = \begin{cases}
    \kappa\lambda_t + \nu \gamma_t & \text{if } \lambda_t\geq0 \text{ and } \kappa\lambda_t+\nu\gamma_t \geq 0\\
    0,              & \text{otherwise}
\end{cases} \quad\\
& \qquad\qquad\qquad t = 0,\mydots,N-1\\
& x_{\textrm{min}} \leq x_t \leq x_{\textrm{max}}, \quad t = 0,\mydots,N \\
& x_0 = x_{\textrm{init}},\\
\end{array}
\end{equation}
where the state $x_t \in \reals^4$ consists of the position of the cart $x_t^1$, angle of the pole $x_t^2$, and their derivatives $x_t^3$ and $x_t^4$, respectively. The force applied to the cart is $u_t\in\reals$ and $s_t\in\reals^2$ are the contact forces imparted by the two walls. The relative distance of the tip of the pole with respect to the left and right walls is $\lambda_t \in \reals^2$ and the time derivative of this relative distance $\gamma_t \in \reals^2$. Finally, $\kappa$ and $\nu$ are parameters associated with the soft contact model used.

As the contact force $s_t$ becomes active only when the tip of the pole makes contact with either wall, we must introduce binary variables to enforce the logical constraints given in \eqref{eq:cart-pole_ocp}. We first define the the relative distance of the tip of the pole with respect to the left and right walls as $\lambda_{t}^1=-x_t^1 + l x_t^2 - d$ and $\lambda_{t}^2 = -x_t^1 - l x_t^2 - d$, respectively. The time derivatives of the relative distance are denoted  $\gamma_{t}^1 = -x_t^3 + l x_t^4$ and $\gamma_{t}^2 = -x_t^3 - l x_t^4$. Using values from $x_{\textrm{min}}$ and $x_{\textrm{max}}$, we can derive explicit upper and lower bounds $\lambda_\textrm{min}^1 \leq \lambda_t^1\leq\lambda_\textrm{max}^1$ and $\lambda_\textrm{min}^2 \leq \lambda_t^2\leq \lambda_\textrm{max}^2$. Similarly, there exist upper and lower bounds $\gamma_\textrm{min}^1 \leq \gamma_t^1 \leq \gamma_\textrm{max}^1$ and $\gamma_\textrm{min}^2 \leq \gamma_t^2 \leq \gamma_\textrm{max}^2$. For further details, we refer the reader to \cite{MarcucciDeitsEtAl2017} for a thorough derivation of the system constraints.

To constrain contact forces $s_t$ to become active only when the pole tip strikes a wall, we introduce four binary variables $\bin_t^i, \quad i = 1,\mydots,4$. Using the formulation from \cite{BemporadMorari1999}, we enforce the following constraints for $k = 1, 2$:
\begin{equation} \label{eq:cartpole_logic}
\begin{array}{l}
\lambda_\textrm{min}^k(1-\bin_t^{(2k-1)}) \leq \lambda_t^k \leq \lambda_\textrm{max}^k\bin_t^{(2k-1)} \nonumber\\
s_\textrm{min}^k(1-\bin_t^{(2k)}) \leq \kappa \lambda_t^k + \nu \gamma_t^k \leq s_\textrm{max}^k\bin_t^{(2k)} \nonumber \\
\end{array}
\end{equation}
Finally, we impose constraints on $s_t$, for $k=1,2$:
\begin{equation} \label{eq:cartpole_contact}
\begin{array}{l}
\nu \gamma_\textrm{max}^{k}(\lambda_t^{k}-1) \leq s_t^{k} - \kappa \lambda_t^{k} - \nu \gamma_t^{k} \leq s_\textrm{min}^k(\bin_t^{2k}-1) \nonumber \\
\end{array}
\end{equation}
There are then a total of $4N$ integer variables in this MIQP. 

\subsubsection{Results}
In this work, the value for $N$ was set to $10$, yielding $40$ total integer variables. The parameter space $\theta \in \reals^{13}$ consists of the initial state $x_0\in\reals^4$, goal state $x_g \in \reals^4$, the relative distance of the tip to the left and right walls $(\lambda_0^1,\lambda_0^2)\in\reals^2$ for the initial state and $(\lambda_g^1,\lambda_g^2)\in\reals^2$ for the goal state, and the distance $||x_0-x_g||_2$. 

Figure \ref{fig:cart-pole_results} shows the results for the cart-pole system over a test set of ten thousand problems. Figure \ref{fig:cart-pole_percent_succ} reports the percent of feasible solutions found over the test set (note that all problems are feasible, so Gurobi reports 100\% here). Among solutions for each method that were feasible, Figure \ref{fig:cart-pole_time} reports the solution time for the forward pass of the network plus computation time for QPs solved before a feasible solution was found. Figures \ref{fig:cart-pole_solved} and \ref{fig:cart-pole_cost} report the number of convex relaxations solved per problem and the cost of the feasible solution relative to the globally optimal solution.

We see that cart-pole system outperforms Gurobi and the regressor benchmarks. For this system, \coco{} finds feasible solutions for 99\% of the problems compared to only 58\% for the regressor. Moreover, we see in Figure \ref{fig:cart-pole_solved} that \coco{} is able to find a feasible solution after one QP solve for 95\% of its feasible problems compared to the average value of 50 QP solves required for Gurobi's branch-and-bound implementation. Note that the regressor always requires one QP solve as only one candidate solution is considered for every forward pass. However, this solution speed-up for \coco{} does not come at the cost of optimality. As shown in Figure \ref{fig:cart-pole_cost}, 99\% of the feasible solutions found by \coco{} are also the globally optimal solution for that problem.

\subsection{Free-Flying Space Robots}

Motion planning in the presence of obstacles is a fundamental problem in robotics \cite{LaValle2006}. Here, we consider a free-flying spacecraft robot in the plane that must navigate through a workspace with multiple obstacles. In this section, we show how using the task-specific strategies approach discussed in \ref{subsec:task-specific_strategies} make this problem tractable to solve with \coco{}.

We denote the position as $p_t \in \reals^2$ and velocity as $v_t$ in the 2-dimensional plane. The state is therefore $x_t = (p_t, v_t)$. The input $u_t\in\reals^2$ consists of the forces produced by the thruster. Letting $\mathcal{X}_{\textrm{safe}}$ be the free space which the robot must navigate through, the optimal control problem is to plan a collision free trajectory towards a goal state $x_g$ \cite{BonalliCauligiEtAl2019}:
\begin{equation} \label{eq:free-flyer_ocp}
\begin{array}{ll}
\text{minimize} & \displaystyle \|x_N-x_g\|_{2} + \sum \limits_{\tau=0}^{N-1} \|x_\tau-x_g\|_{2} + \|u_\tau\|_{2} \\
\text{subject to} & x_{t+1} = Ax_t + Bu_t, \!\! \quad t = 0,\mydots,N-1 \\
& ||u_t||_{2} \leq u_{\textrm{max}},\!\! \quad t = 0,\mydots,N-1 \\
& x_{\textrm{min}} \leq x_t \leq x_{\textrm{max}},\!\! \quad t = 0,\mydots,N \\
& x_0 = x_{\textrm{init}}\\
& x_t \in \mathcal{X}_{\textrm{safe}},\!\! \quad t=0,\mydots,N,
\end{array}
\end{equation}
Here, the crucial constraint distinguishing \eqref{eq:free-flyer_ocp} from a typical optimal control problem is the safety constraint $x_t \in \mathcal{X}_{\textrm{safe}}$. In the presence of obstacles, this constraint is typically highly non-convex and requires a global combinatorial search to solve this problem. One class of methods used to solve \eqref{eq:free-flyer_ocp} entails posing it as an MICP and using binary variables to enforce collision-avoidance constraints~\cite{MoteEgerstedtEtAl2020,SchouwenaarsDeMoorEtAl2001}. Given a set of $N_\textrm{obs}$ obstacles, the workspace is decomposed into keep-in and keep-out polytope regions, where binary variables are then used to enforce collision avoidance with the keep-out regions.

To simplify the presentation, we consider axis-aligned rectangular obstacles, but the framework can be generalized to any obstacles represented as convex polygons~\cite{LandryDeitsEtAl2016}.
We represent an obstacle $m$ with the coordinates of the rectangle for the lower-left hand corner $(x_{\textrm{min}}^m,y_{\textrm{min}}^m)$ and upper right-hand corner $(x_{\textrm{max}}^m,y_{\textrm{max}}^m)$.
Given the state $x_t$, the collision avoidance constraints with respect to obstacle $m$ are:
\begin{align}
x_{\textrm{max}}^m + M\bin_{t}^{m,1} \leq x_t^1 \leq x_{\textrm{min}}^m + M\bin_{t}^{m,2} \label{eq:obstacle_avoidance_x}\\
y_{\textrm{max}}^m + M\bin_{t}^{m,3} \leq x_t^2 \leq y_{\textrm{min}}^m + M\bin_{t}^{m,4} \label{eq:obstacle_avoidance_y}
\end{align}

\begin{figure*}[tb!]
\centering
  \begin{subfigure}[t]{0.2\textwidth}
  \centering
  \captionsetup{justification=centering}
    \includegraphics[width=1.75in,]{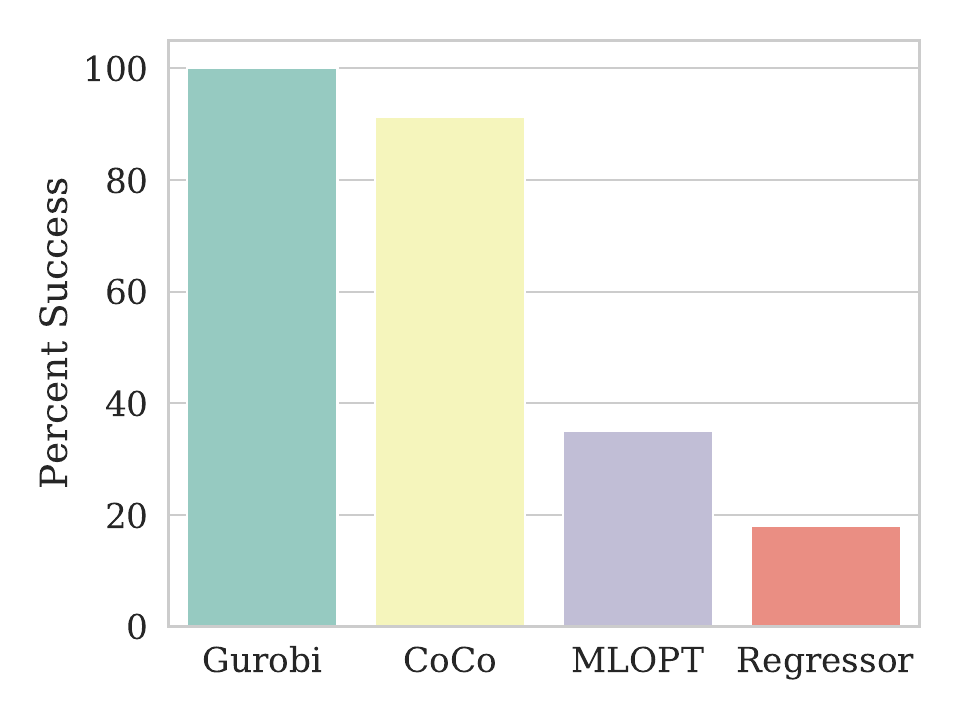}
    \caption{Success percentage}
    \label{fig:free-flyer_percent_succ}
  \end{subfigure}
  \qquad
  \begin{subfigure}[t]{0.2\textwidth}
  \centering
  \captionsetup{justification=centering}
    \includegraphics[width=1.75in]{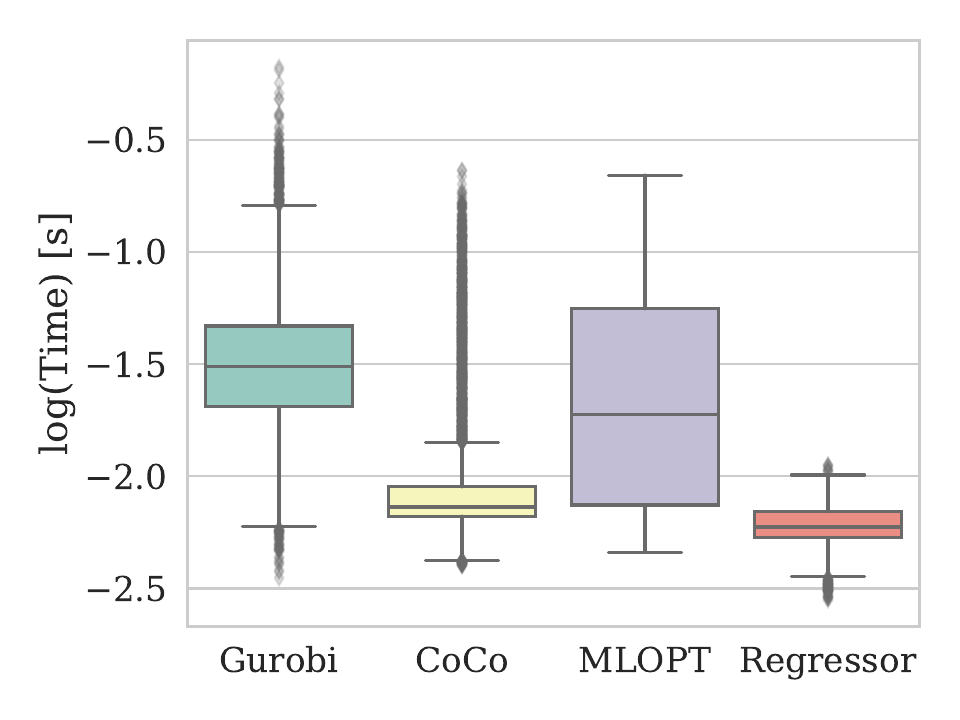}
    \caption{Solution times}
    \label{fig:free-flyer_time}
  \end{subfigure}
  \qquad
  \begin{subfigure}[t]{0.2\textwidth}
    \centering
    \captionsetup{justification=centering}
    \includegraphics[width=1.75in]{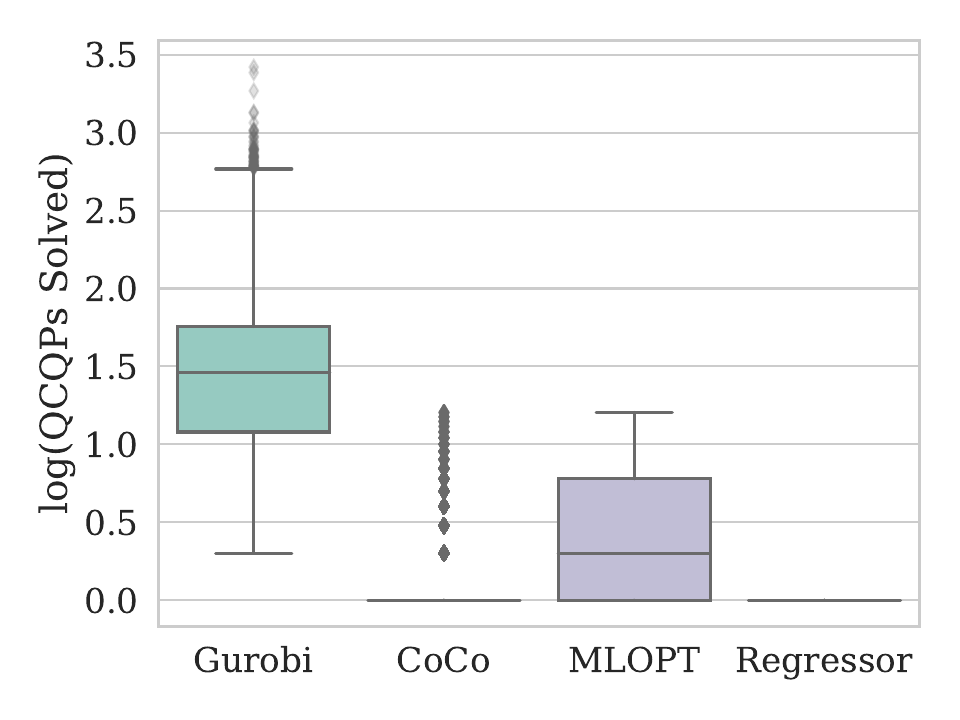}
    \caption{Num. QCQPs solved}
    \label{fig:free-flyer_solved}
  \end{subfigure}
  \qquad
    \begin{subfigure}[t]{0.2\textwidth}
    \centering
    \captionsetup{justification=centering}
    \includegraphics[width=1.75in]{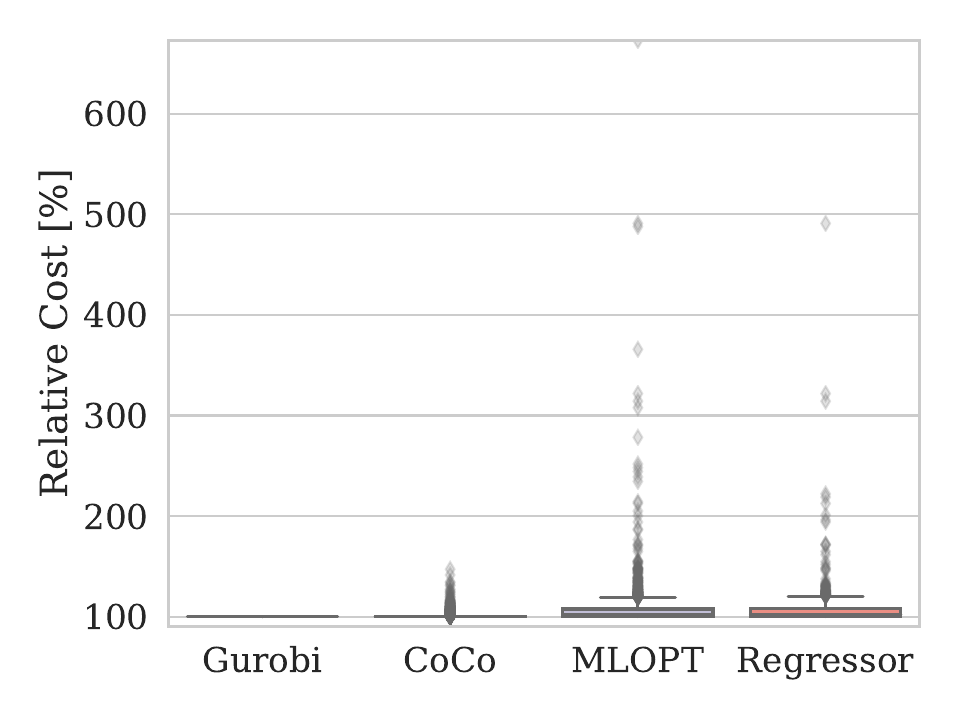}
    \caption{Normalized cost [\%]}
    \label{fig:free-flyer_cost}
  \end{subfigure}
  \caption{Simulation results for the free-flyer show how task-specific strategies are critical for enabling the use of \coco{} for this system.}
\label{fig:free-flyer_results}
\end{figure*}

For an obstacle set of cardinality $N_\textrm{obs}$, each obstacle thus requires four integers variables $\bin_{t}^{m,i}$ at each time step. Because the robot must be in violation of at least one of the keep-out constraints, $\bin_t^{m,i}=1$ corresponds to being on one side of face $i$ of the rectangle. A final constraint is enforced to ensure that the robot does not collide with the obstacle:
\begin{equation}
\begin{array}{l}
\sum\limits_{i=1}^4 \bin_{t}^{m,i} \leq 3, \quad m = 1, \mydots, N_{\textrm{obs}},  \quad t = 1, \mydots, N. \label{eq:obstacle_avoidance_sum}
\end{array}
\end{equation}
Due to the $\ell_2$-norm constraints imposed on the thruster forces $u_t$, this problem is an MIQCQP with $4N_{\mathrm{obs}}N$ variables.

\subsubsection{Task-Specific Strategy Decomposition}
We show here how the the underlying structure of the obstacle avoidance constraints can be leveraged for effectively training a strategy classifier, as discussed in~\ref{subsec:task-specific_strategies}. We note that each binary variable depends only on the three other variables associated with the same obstacle at the same timestep. Thus, we can decompose the strategy on a per-obstacle basis. Thus, each strategy $\mathcal{S}(\theta;m)$ is comprised of a tuple $(\bin^{m,*}, \mathcal{T}_M^m)$, where $\bin^{m,*}$ and $\mathcal{T}_M^m$ are defined as in \eqref{eq:strategy_defn} for only the integer assignments and big-M constraints for obstacle $m$.

Rather than training $N_{\textrm{obs}}$ separate strategy classifiers for each $\mathcal{S}(\theta;m)$, we train a single classifier using $\theta$ and a one-hot vector to encode which obstacle strategy is being queried.

\subsubsection{Results}
 In this work, the horizon $N$ was set to $5$ and the number of obstacles $N_{\mathrm{obs}}$ to $8$, yielding $160$ total integer variables. The parameter space $\theta \in \reals^{44}$ for this problem included the initial condition $x_0\in\reals^4$ and coordinates $(x_\textrm{min}^m,x_\textrm{max}^m,y_\textrm{min}^m,y_\textrm{max}^m)\in\reals^4$ of each of the eight obstacles $m=1,\ldots,8$, and a one-hot encoding of the obstacle index.

 Figure \ref{fig:free-flyer_results} shows the results over the ten thousand test problems. The trained \coco{} classifier is compared against Gurobi and the regressor. We also benchmark against an implementation of \mlopt{} that does not decompose the strategies over each obstacle separately. The number of integer strategies observed in the training set for \mlopt{} was approximately 81 thousand, whereas the task-specific strategy decomposition for \coco{} resulted in 516 strategies. As shown in Figure \ref{fig:free-flyer_percent_succ}, \coco{} finds feasible solutions for 92\% of the training set compared to 18\% for the regressor. Crucially, we see that the \mlopt{} finds feasible solutions for only 35\% of the test set. \coco{} is on average also able to find solutions faster than Gurobi and \mlopt{}. Figure \ref{fig:free-flyer_solved} shows that this faster computation time is achieved by \coco{} having to solve only one QCQP for 85\% of the test problems. Once again, \coco{} is still able to find high quality feasible solutions and finds the optimal solution for 92\% of the time as illustrated in Figure \ref{fig:free-flyer_cost}.

For this system, we see that the task-specific strategy approach was required in enabling \coco{} to reliably find solutions for the MIQCQPs. \mlopt{} finds feasible solutions for 56\% fewer problems than the task-specific \coco{}. This gap is likely attributable to the fact that the number of integer strategies \mlopt{}  has to consider is approximately 81 thousand compared to 516 for the task-specific \coco{}. Thus, the task-specific strategies also allow for better supervision of the classifier as there are more labels available per class.

\subsection{Task-Oriented Optimization of Dexterous Grasps}
As a final system, we consider the problem of grasp optimization for task-specific dexterous grasps. Practically, during dexterous manipulation (especially with environmental contact) the actual trajectory of the object can diverge significantly from the planned trajectory; further, high-level tasks (such as placing a peg into a hole) may require multiple grasps for various subtasks. Thus, we are interested in enabling online computation of optimal dexterous grasps for both fast replanning and regrasping.

While task-agnostic grasp optimization has been well-studied \cite{FerrariCanny1992}, the problem choosing and even evaluating dexterous grasps for specific tasks (such as pushing or rotating objects, tool use) is still of considerable interest. While works such as \cite{LiSastry1988,HaschkeSteilEtAl2005} have proposed metrics which can be used to evaluate grasps for various tasks, they leave the problem choosing grasps which optimize these metrics unstudied.



\begin{figure}[hb]
\centering
\includegraphics[scale=0.25,]{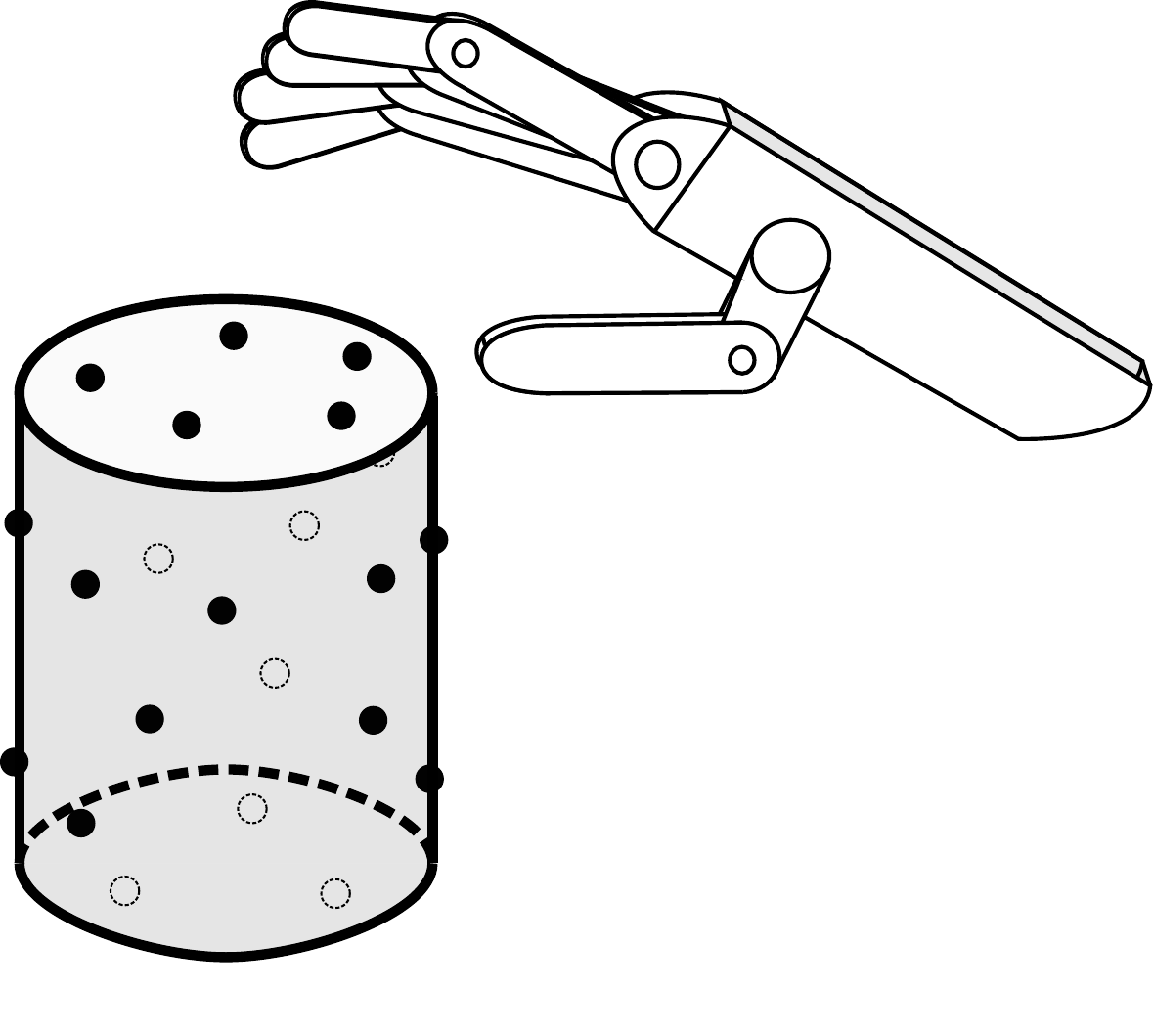}
\caption{Schematic of dexterous grasping problem. Here, a robotic hand with $n = 5$ fingers chooses from $M$ potential contact points to optimize a task-specific grasp metric. }
\label{fig:task-specific_grasping}
\end{figure}

Here we consider the problem of choosing $n$ contact points for a multifingered robot hand from a set of points $p_1, \mydots, p_M \in \mathbf{R}^3$, sampled from the object surface in order to optimize the task-oriented grasp metric proposed in \cite{HaschkeSteilEtAl2005}. Figure \ref{fig:task-specific_grasping} shows a schematic of the proposed problem. We model the contacts between the fingers and the object as point contacts with friction. Thus, at each candidate point, if selected, a finger could apply a local contact force $f_i = (f_i^x, f_i^y, f_i^z)\in \reals^3$, where the local coordinate frame has the $x$- and $y$-axes tangent to the surface, and the $z$-axis along the inward surface normal. Intuitively, $f_i^z$ is the component of the contact force which is normal to the object surface, and $f_i^x, f_i^y$ are its tangential components. 


Under this contact model, the contact force $f^{(i)}$ must lie in the \textit{friction cone} $\mathcal{K}^{(i)}$, which constrains the tangential component of the contact force to be less than the friction coefficient $\mu_i$ times the normal force.
We can further define the \textit{grasp matrices} $G_1, \ldots, G_M$, where
\begin{equation*}
    G_i = \left[\begin{array}{c}R^{(i)}\\ \left(p^{(i)}\right)^\times R^{(i)} \end{array}\right],
\end{equation*}
$R^{(i)} \in SO(3)$ is the rotation matrix relating the local frame of point $i$ to the global frame, and, for a vector $v \in \reals^3$, $(v)^\times$
\begin{equation*}
(v)^\times = \left[\begin{array}{ccc} 0 &-v_3 &v_2\\ v_3 &0 &-v_1\\ -v_2 &v_1 &0\end{array}\right]
\end{equation*}
is the skew-symmetric matrix such that for $w \in \reals^3$, $$(v)^\times w = v \times w.$$ We can now express the wrench applied to the object by $f^{(i)}$ as $G_i f^{(i)}$, and the total wrench (from all contact forces) as $G f$, where $G = \left[G_1, \ldots, G_M\right]$.

However, contact forces may not be applied at all candidate points. To this end, we introduce the logical variables $\bin_i \in \left\{0,1\right\}$, with $\bin_i= 1$ iff point $p_i$ is selected for the grasp. Thus, we enforce the constraint $$f_i^z \leq \bin_i,$$ which constrains the normal forces of all unused grasps to be zero, and to be bounded by unity otherwise. Thus, for a choice of grasps $\bin = (\bin_1, \mydots, \bin_M)$, the set of possible object wrenches is defined as $$\mathcal{W}(\bin) = \left\{G f \mid f\in \mathcal{K}_i, f_i^z \leq \bin_i \right\}.$$

A common task-agnostic metric for grasp quality is the radius $\alpha$ of the largest ball which can be inscribed in $\mathcal{W}(\bin),$ $$\mu_g(\bin) = \sup\left\{\alpha \geq 0 \mid B_\alpha \subset \mathcal{W}(\bin)\right\},$$
where $B_\alpha$ denotes the ball centered at the origin of radius $\alpha$. Intuitively, this gives the norm of the smallest applied wrench, in any direction, that lies on the boundary of $\mathcal{W}(\bin)$, which we denote as $\partial \mathcal{W}$. This can be understood as a measure of the grasp's general ``control authority'' over the body. Of course, few tasks require equal wrenches to be generated in every direction, so for many tasks this metric is over-conservative; optimal grasps for this metric are usually enveloping grasps which may be inappropriate for tasks such as pushing.

\begin{figure}[hb]
\centering
\includegraphics[scale=0.4,]{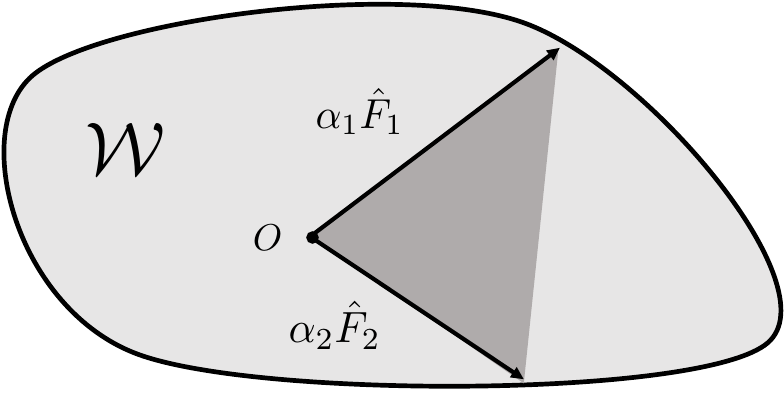}
\caption{Schematic of task space $\mathcal{W}$ for a specific grasp $\delta$, along with the task vectors $\hat{F}_t$, their qualities $\alpha_t$, and the task polyhedron (shown in dark gray). The posed problem corresponds to choosing $\delta^*$ which maximizes the (weighted) volume of the task polyhedron.}
\label{fig:task-specific_grasping}
\end{figure}

In contrast, in \cite{HaschkeSteilEtAl2005}, the authors propose describing tasks using normalized \textit{task wrenches} $\hat{F}_t$, which are specific directions in wrench space that characterize the applied wrenches necessary to complete the task. For instance, if the desired task is to push the object along the $+x$-axis, then this task could be described using $\hat{F} = (1,0,0,0,0,0)$, and so on. Thus, for a task described by a single wrench, the grasp quality can be defined as $$\mu_1(\bin, \hat{F}_t) = \sup \left\{\alpha_t \geq 0 \mid  \alpha_t \hat{F}_t \in \partial \mathcal{W}(\bin) \right\}.$$
For a given grasp $\bin$, this metric can be easily computed by solving a second-order cone program (SOCP).

However, most tasks are best described by a set of wrenches which must be generated, rather than a single direction in wrench space. Thus, the authors propose describing this set as the positive span of $T$ task vectors; in turn, the grasp metric is defined as $$\mu(\bin, \hat{F}_1, \ldots, \hat{F}_T) = \textstyle \sum \limits_{t=1}^{T} w_t \alpha_t,$$ where $w_i \geq 0$ are the relative weightings of the task vectors, and $\alpha_t = \mu_1(\bin,\hat{F}_t).$ This can, in turn, be computed by solving $T$ SOCPs. 
This corresponds to the volume of the polyhedron defined by the vectors $w_t \alpha_t \hat{F}_t$.

We seek $\bin^*$ which maximizes this grasp metric, which can be written as a mixed-integer SOCP (MISOCP):
\begin{equation} \label{eq:manip_prob}
\begin{array}{ll}
\text{maximize} & \displaystyle \sum\limits_{t=1}^{T} w_t \alpha_t\\
\text{subject to} &G f^t = \alpha_t \hat{F}_t , \quad t = 1, \ldots, T\\
 &f_i^t \in \mathcal{K}^{(i)}, \quad i = 1, \ldots M,\; t = 1, \ldots, T\\
 & f_i^{z,t} \leq \bin_i, \quad i = 1, \ldots, M,\; t = 1, \ldots, T\\
 & \sum\limits_{i=1}^M \bin_i \leq n\\
 & \bin \in \left\{0,1\right\}^M
\end{array}
\end{equation}

\begin{figure*}[tb!]
\centering
  \begin{subfigure}[t]{0.2\textwidth}
    \centering
  \captionsetup{justification=centering}
    \includegraphics[width=1.75in,]{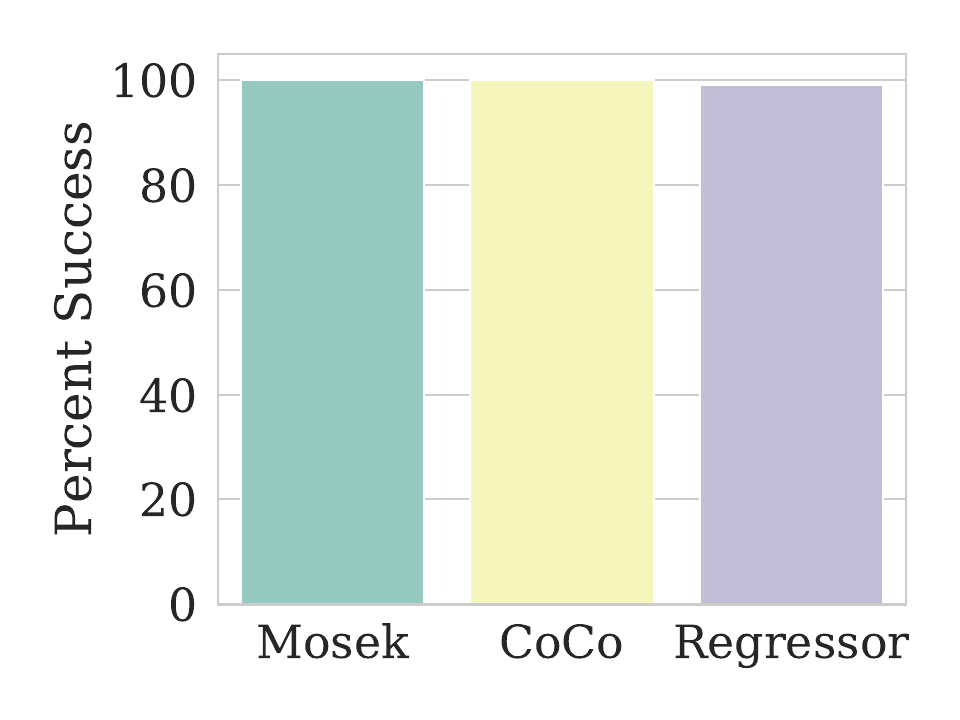}
    \caption{Success percentage}
    \label{fig:manipulation_percent_succ}
  \end{subfigure}
  \qquad
    \begin{subfigure}[t]{0.2\textwidth}
    \centering
  \captionsetup{justification=centering}
    \includegraphics[width=1.75in,]{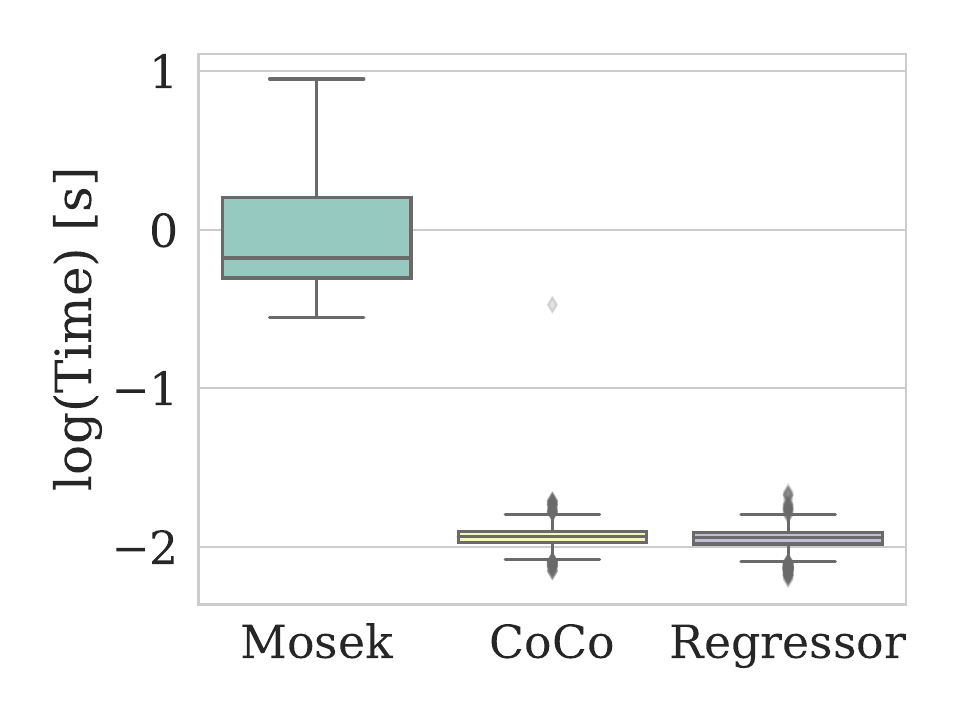}
    \caption{Solution times}
    \label{fig:manipulation_time_succ}
  \end{subfigure}
  \qquad
    \begin{subfigure}[t]{0.2\textwidth}
    \centering
  \captionsetup{justification=centering}
    \includegraphics[width=1.75in,]{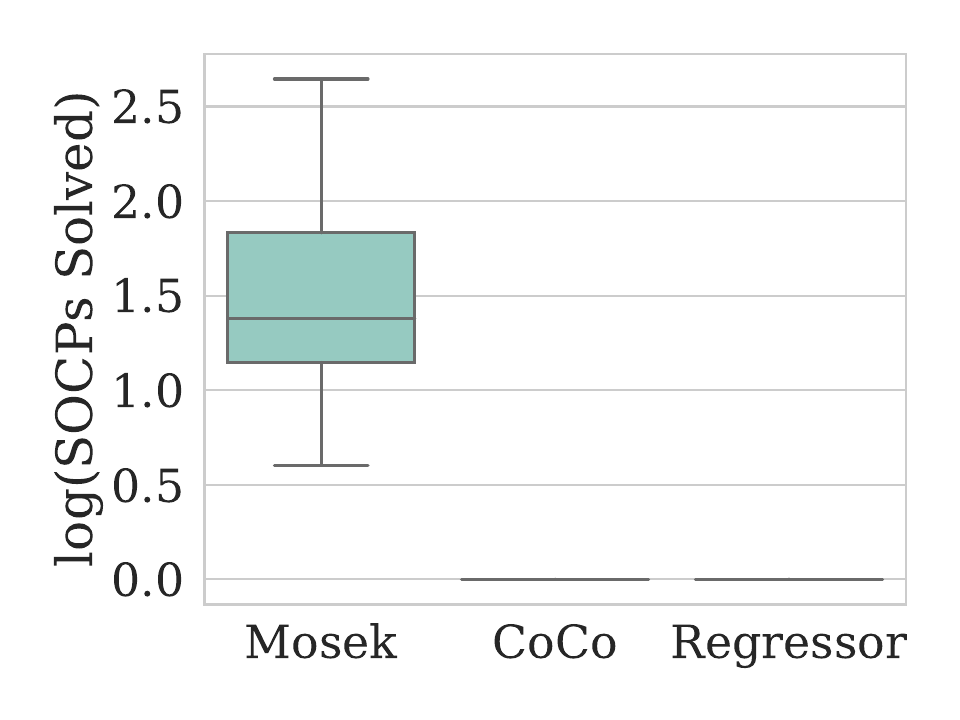}
    \caption{Num. SOCPs solved}
    \label{fig:manipulation_solved}
  \end{subfigure}
  \qquad
      \begin{subfigure}[t]{0.2\textwidth}
    \centering
  \captionsetup{justification=centering}
    \includegraphics[width=1.75in,]{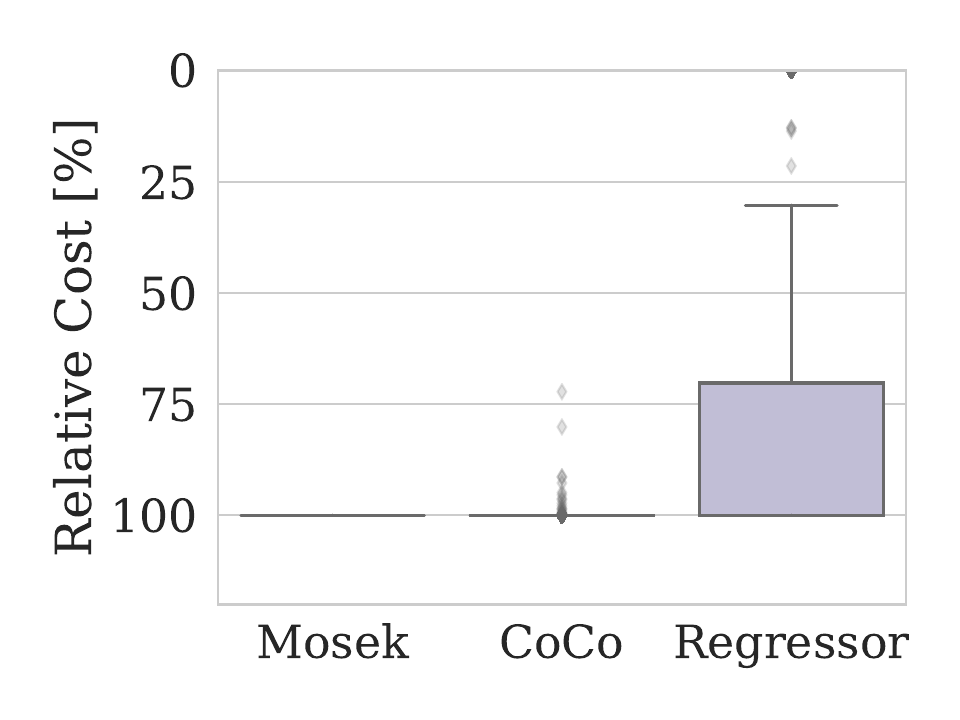}
    \caption{Normalized cost [\%]}
    \label{fig:manipulation_cost}
  \end{subfigure}
  \caption{Simulation results for manipulation example. \coco{} reduces solution times for \eqref{eq:manip_prob} between 2--3 orders of magnitude.}
\label{fig:manipulation_results}
\end{figure*}

\begin{figure}[t!]
\centering
\includegraphics[width=0.8\columnwidth]{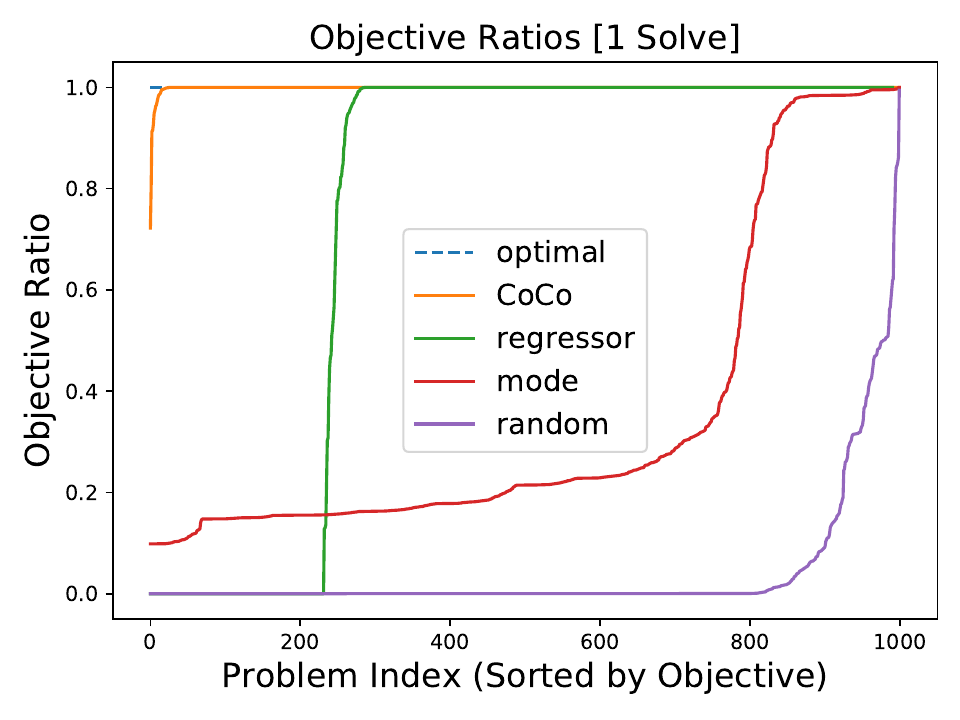}
\caption{The optimality gap plotted across 1000 problems in the training dataset for grasp modes chosen by \coco{}, regressor, the most common mode seen in the training data, and at random.}
\label{fig:manipulation_cost_comparison}
\end{figure}

\subsubsection{Results}
For our numerical experiments, we aimed to learn strategies for a single rigid body and set of candidate points which generalized across task weightings $w$. We set the number of candidate grasp point $M$ equal to $30$ and plan grasps for a four finger manipulator $n=4$.

For training, we collected a set of 4,500 problems, using task vectors $\hat{F}_t$ which corresponded to the basis vectors $\pm e_i \in \reals^6$ for $i = 1, \ldots, 6$. Since these task vectors span $\reals^6$, task weightings $w$ with all $w_i > 0$ correspond to generating force closure grasps. We generated task weightings $w_i$ by calculating the softmax of a vector sampled from a multivariate normal distribution with covariance matrix $\Sigma = 10\mathbf{I}$.

Figures \ref{fig:manipulation_results} and \ref{fig:manipulation_cost_comparison} show the results for this system. Note that because $\alpha_t$ can be set to 0 \ie, resist a zero wrench, all grasp mode sequences with four contacts are feasible. However, we see in Figure \ref{fig:manipulation_percent_succ} that the regressor can report infeasible solutions if it guesses more than four active binary assignments. Although all \coco{} candidate strategies are feasible, we see in Figure \ref{fig:manipulation_cost_comparison} that maximizing the grasp metric is nonetheless challenging for the regressor and simple heuristics. Thus, Figure \ref{fig:manipulation_cost} illustrates that the feasible solution found by \coco{} is also the globally optimal grasp for 99\% of the problems (note that the scale is inverted in this figure as the MISOCP is a maximization problem).

\begin{figure*}[tb!]
\centering
  \begin{subfigure}[t]{0.3\textwidth}
    \centering
  \captionsetup{justification=centering}
    \includegraphics[width=\textwidth]{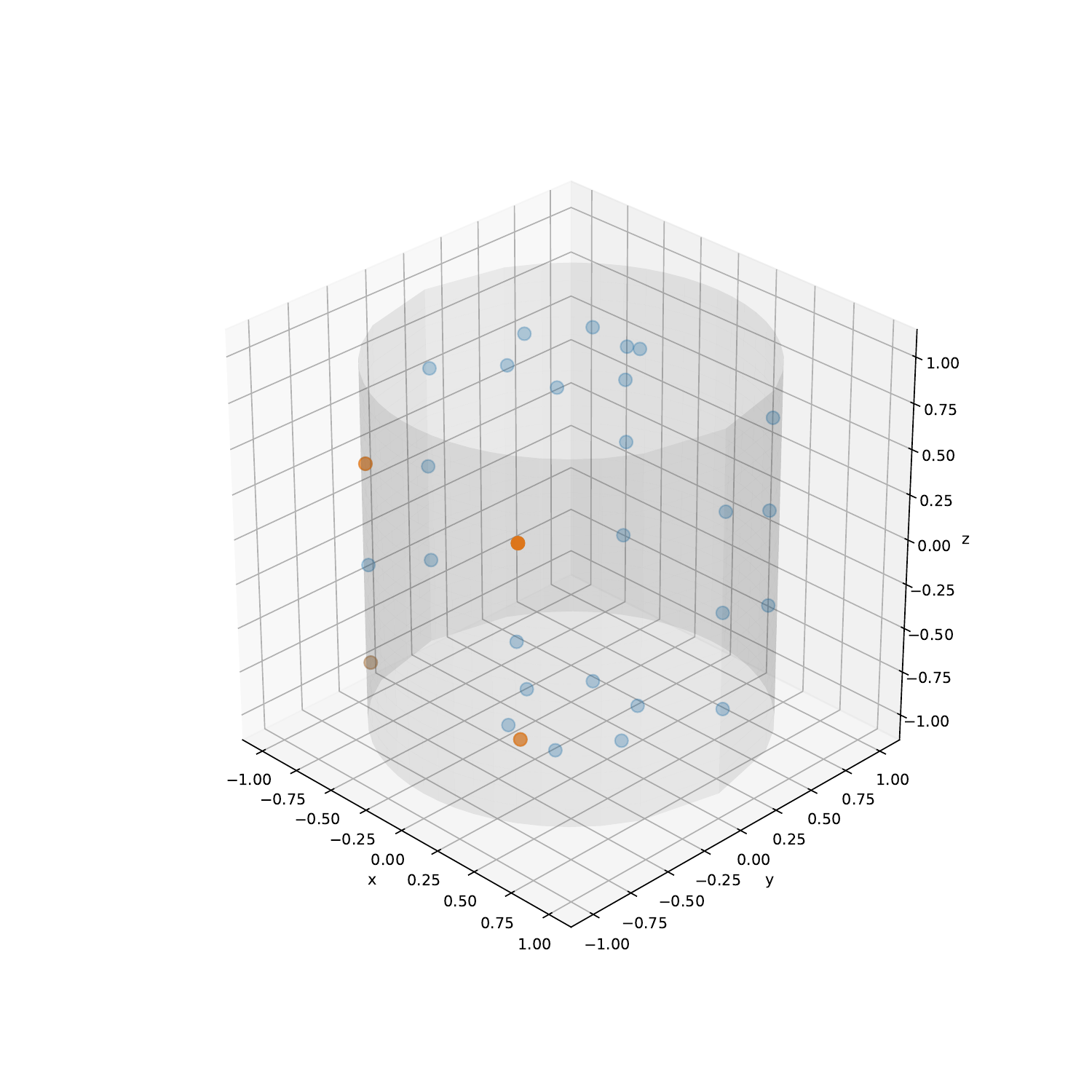}
    \caption{Force in $+y$ direction}
    \label{fig:manipulation_visualization_+y}
  \end{subfigure}
  \qquad
    \begin{subfigure}[t]{0.3\textwidth}
    \centering
  \captionsetup{justification=centering}
    \includegraphics[width=\textwidth]{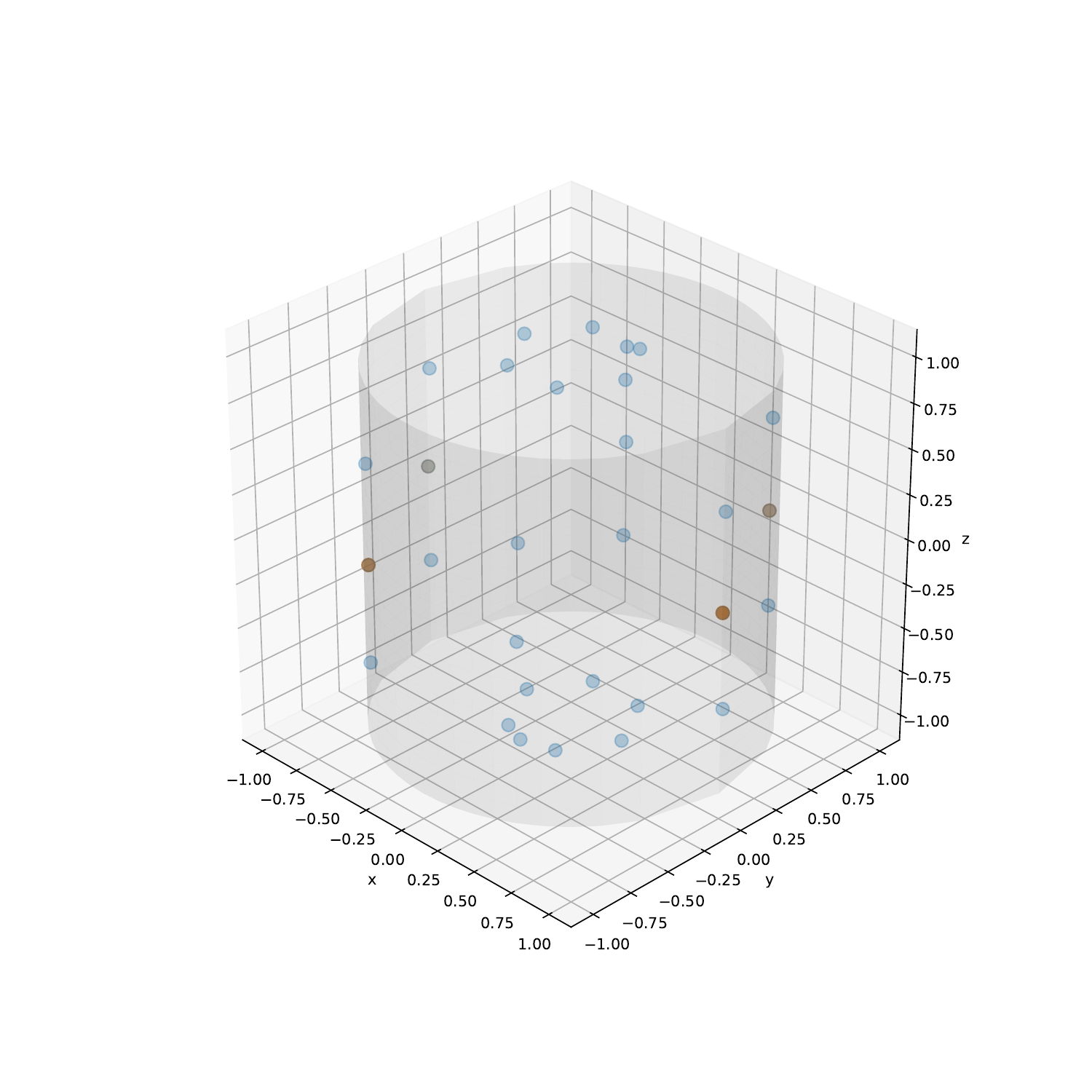}
    \caption{Moment in $+z$ direction}
    \label{fig:manipulation_visualization+Mz}
  \end{subfigure}
  \qquad
    \begin{subfigure}[t]{0.3\textwidth}
    \centering
  \captionsetup{justification=centering}
    \includegraphics[width=\textwidth]{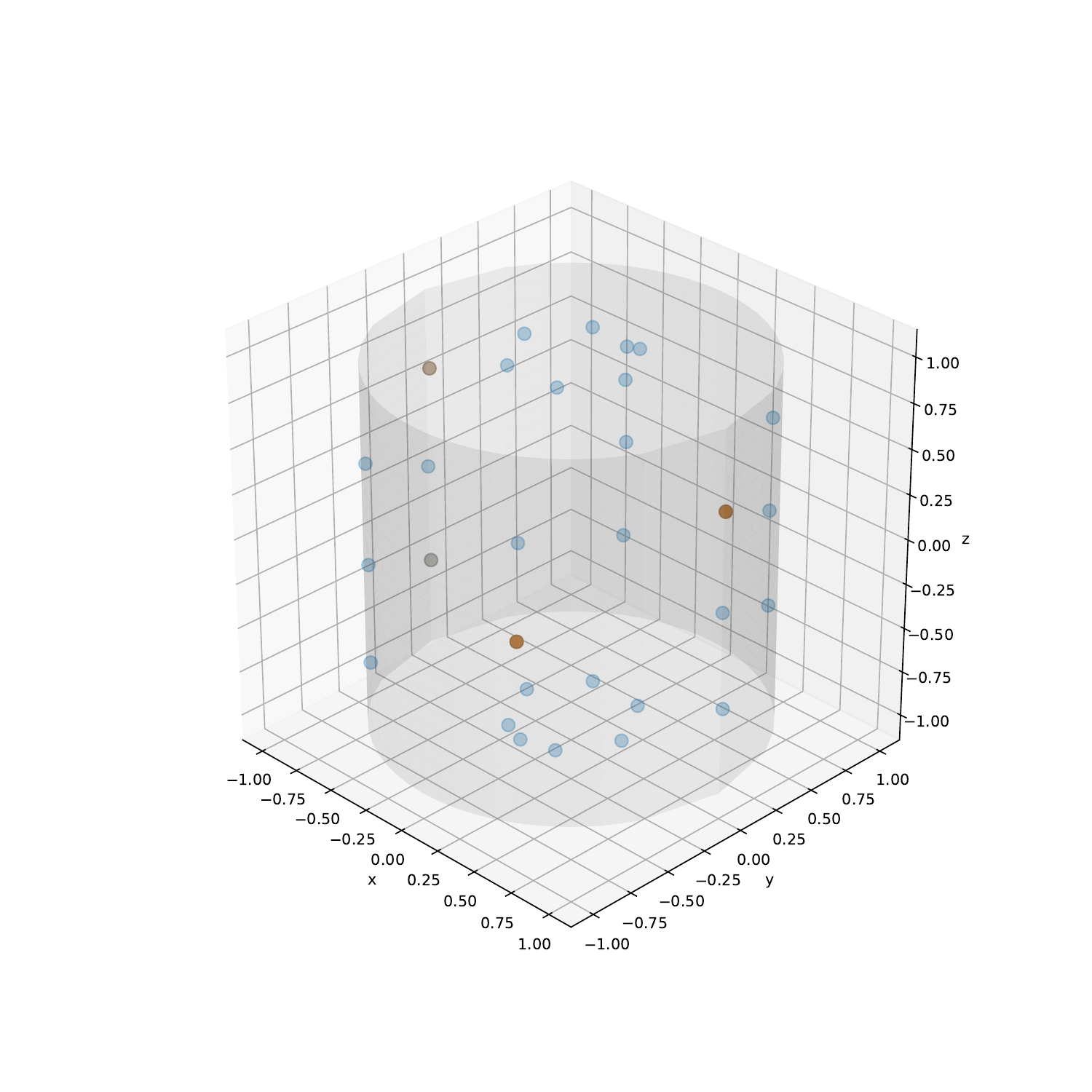}
    \caption{Volume maximization, $w_i = \frac{1}{12}$}
    \label{fig:manipulation_visualization_force_closure}
  \end{subfigure}
  \caption{Visualization of grasps selected by network policy for various task weightings. The candidate points are shown in blue, with the selected points shown in red. We note choosing a weighting $w_i = \frac{1}{12}$ for all $i$, which approximates the equally-weighted ellipsoid proposed in \cite{LiSastry1988}, results in a reasonable ``enveloping'' grasp, as shown in (c).}
\label{fig:manipulation_visualization}
\end{figure*}

Figure \ref{fig:manipulation_visualization} shows the grasps selected by \coco{} for some representative task weightings. Intuitively, the highest-scoring grasp when all weight is placed on creating force in the $+y$ direction uses the four points with minimal $y$ values. Similarly, the top-scoring grasp for creating a moment about the $+z$-axis uses points on the radius of the cylinder, maximizing their moment arm from the center of mass. Finally, in the case of equal weighting for all directions, which approximates the unweighted grasp ellipsoid studied in \cite{LiSastry1988}, the top-scoring grasp qualitatively resembles an ``enveloping'' grasp, selecting points which are somewhat evenly distributed about the cylinder.

Crucially, this ability to find the globally optimal solution for the majority of problems after solving only one SOCP relaxation leads to solutions in tens of milliseconds rather than the seconds needed by Mosek. On average, \coco{} is able to accelerate finding solutions for this problem by two to three orders of magnitude, making solving MISOCPs in real-time with high-quality solutions tractable.



\section{Conclusion}
\label{sec:Conclusion}
In this paper, we presented \coco{}, a machine learning technique to accelerate the solution of online MICPs by exploiting the structure of problems arising in practical applications. We discarded redundant strategies corresponding to equivalent globally optimal solutions. In addition, we exploited the separable structure of robotics problems to design task-specific strategies. Numerical examples show that our approach results in greater than $90\%$ feasibility, with more than $90\%$ of solutions being globally optimal in common robotics setups such as a cart-pole system with walls, a free-flying space robot, and task-oriented grasps. We also obtained 1-2 orders of magnitude speedups compared to commercial solvers. The proposed algorithm is, therefore, suitable to compute MICP solutions in real-time with high reliability and speed.

Future contributions will focus on providing theoretical guarantees on how to characterize the strategy space by effectively sampling the parameter space of the problems and to bound the optimality gap of feasible solutions found online. Additionally, we would like to explore the application of this proposed framework towards mixed-integer non-convex problems appearing in robotics such as task planning.

\section*{Acknowledgements}
We would like to thank Matteo Zallio, Benoit Landry, and Joseph Lorenzetti for their discussions during this work.

{\small
\renewcommand{\baselinestretch}{0.96}
\bibliographystyle{IEEEtran}
\bibliography{ASL_papers,main}}
%




\end{document}